\documentclass[11pt]{article}
\usepackage{microtype} % microtypography
\usepackage{booktabs}  % tables
\usepackage{url}  % urls

\usepackage{amsmath}
\usepackage{amsthm}

\usepackage{multirow}
\usepackage[table,xcdraw]{xcolor}
\usepackage[normalem]{ulem}
\useunder{\uline}{\ul}{}
\usepackage{amssymb}% http://ctan.org/pkg/amssymb
\usepackage{pifont}% http://ctan.org/pkg/pifont
\newcommand{\cmark}{\ding{51}}%
\newcommand{\xmark}{\ding{55}}%
\usepackage{graphicx}
\usepackage{tablefootnote}

\usepackage{listings}
\usepackage{xcolor}

\definecolor{codegreen}{rgb}{0,0.6,0}
\definecolor{codegray}{rgb}{0.5,0.5,0.5}
\definecolor{codepurple}{rgb}{0.58,0,0.82}
\definecolor{backcolour}{rgb}{0.95,0.95,0.92}

\lstdefinestyle{mystyle}{
    backgroundcolor=\color{backcolour},   
    commentstyle=\color{codegreen},
    keywordstyle=\color{magenta},
    numberstyle=\tiny\color{codegray},
    stringstyle=\color{codepurple},
    basicstyle=\ttfamily\footnotesize,
    breakatwhitespace=false,         
    breaklines=true,                 
    captionpos=b,                    
    keepspaces=true,                 
    numbers=left,                    
    numbersep=5pt,                  
    showspaces=false,                
    showstringspaces=false,
    showtabs=false,                  
    tabsize=2
}

\usepackage[final]{automl}
\usepackage[authoryear]{natbib}
\title{AutoGluon-Multimodal (AutoMM): Supercharging Multimodal AutoML with Foundation Models}

\author[1,$\ast$]{\nameemail{Zhiqiang Tang}{zqtang@amazon.com}}
\author[1]{\nameemail{Haoyang Fang}{haoyfang@amazon.com}}
\author[1]{\nameemail{Su Zhou}{zhousu@amazon.com}}
\author[1]{\nameemail{Taojiannan Yang}{taoyaang@amazon.com}}
\author[1]{\nameemail{Zihan Zhong}{zihanzh@amazon.com}}
\author[1]{\nameemail{Tony Hu}{tonyhu@amazon.com}}
\author[1]{\nameemail{Katrin Kirchhoff}{katrinki@amazon.com}}
\author[1]{\nameemail{George Karypis}{gkarypis@amazon.com}}

\affil[1]{Amazon Web Services}
\affil[$\ast$]{ Correspondence to: Zhiqiang Tang <zqtang@amazon.com>}

\hypersetup{%
  pdfauthor={}, % will be reset to "Anonymous" unless the "final" package option is given
  pdftitle={},
  pdfsubject={},
  pdfkeywords={}
}

\begin{document}

\maketitle

\begin{abstract}
AutoGluon-Multimodal (AutoMM) is introduced as an open-source AutoML library designed specifically for multimodal learning.\footnote{\url{https://github.com/autogluon/autogluon}} Distinguished by its exceptional ease of use, AutoMM enables fine-tuning of foundation models with just three lines of code. Supporting various modalities including image, text, and tabular data, both independently and in combination, the library offers a comprehensive suite of functionalities spanning classification, regression, object detection, semantic matching, and image segmentation. Experiments across diverse datasets and tasks showcases AutoMM's superior performance in basic classification and regression tasks compared to existing AutoML tools, while also demonstrating competitive results in advanced tasks, aligning with specialized toolboxes designed for such purposes.
\end{abstract}

% ==== Formatting Instructions
% The page limit for the main paper is 9 pages; this includes the broader impact
% statement but not the submission checklist, references, or appendix.
% The broader impact statement and submission checklist are mandatory at both
% submission time and in the camera ready. References and supplemental materials
% are not limited in length. Accepted papers will be allowed to add an additional page
% of content to the main paper to react to reviewer feedback.
% This additional content may in fact be added during the rebuttal phase as authors
% interact with the reviewers to ensure acceptance decisions can be made regarding
% near-camera-ready work.

\section{Introduction}
Automated machine learning (AutoML) \citep{yao2018taking, zoller2021benchmark, he2021automl} promises to streamline the process of translating raw data into accurate predictions, minimizing the need for extensive human intervention and expertise, though a noticeable gap still exists in some domains \citep{schmarje2021survey, parisi2019continual, chan2020deep}. By encapsulating best practices in machine learning—from data preprocessing \citep{gada2021automated} to model selection \citep{arango2023quick}, training \citep{falcon2019pytorch}, and deployment \citep{paleyes2022challenges}—AutoML frameworks aim to democratize machine learning capabilities, enabling both technical and non-technical users to develop high-performing models efficiently. This scalability of expertise empowers practitioners to tackle a wide array of tasks without requiring deep knowledge of machine learning techniques.

The continual evolution of machine learning techniques, particularly the advent of foundation models \citep{bommasani2021opportunities} that are pre-trained on large-scale datasets and are applicable to a wide array of downstream tasks have revolutionized fields such as computer vision \citep{dosovitskiy2020image, liu2021swin} and natural language processing \citep{devlin2018bert, liu2019roberta}. Fine-tuning these models for specific domains is crucial for extending their utility to end-users, yet dedicated open-source AutoML toolboxes for this purpose remain scarce.
In addition, existing well-known open-source AutoML toolboxes predominantly focus on basic classification and regression tasks with tabular data \citep{thornton2013auto, feurer2015efficient, olson2016tpot, erickson2020autogluon, ledell2020h2o, zimmer2021auto}, overlooking the complexities of real-world problems that often entail multiple modalities \citep{baltruvsaitis2018multimodal}. For instance, tasks like webpage analysis typically involve processing image, text, and tabular data concurrently, with objectives ranging from object detection \citep{zou2023object} to semantic matching \citep{reimers2019sentence}. Despite the availability of specialized tools for individual tasks, there is a notable gap in the AutoML community regarding unified frameworks capable of handling diverse modalities and tasks seamlessly.

%% , underscores the need to advance AutoML methodologies. Foundation models 
%% offering state-of-the-art performance across diverse tasks. 

\begin{figure}[t]
    \centering
    \includegraphics[width=0.9\textwidth,bb=0 0 880 336]{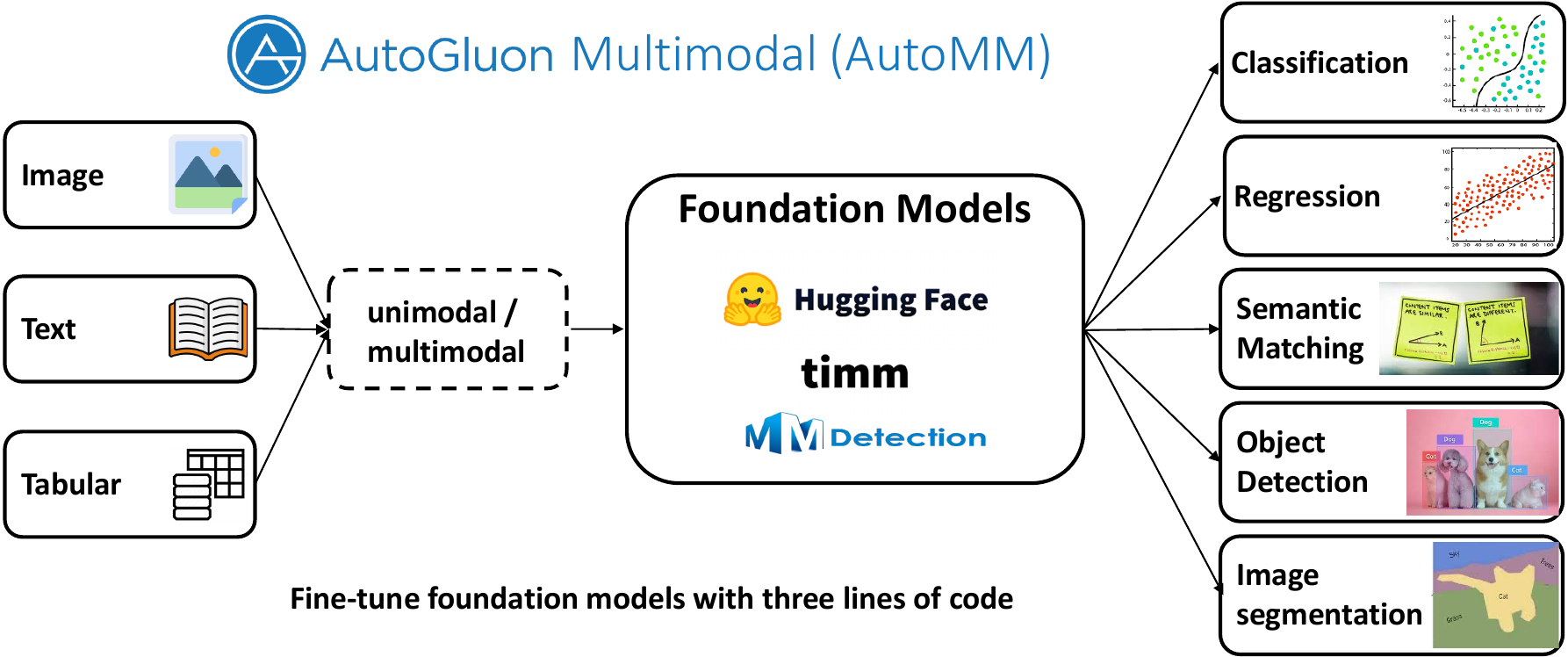}
    \caption{AutoMM introduction: Supporting both unimodal and multimodal data (\textbf{Left}), AutoMM enables seamless fine-tuning of foundation models (\textbf{Middle}) for basic classification/regression as well as advanced tasks (\textbf{Right}).} 
    \label{fig:automm-intro}
\end{figure}

To address these dual challenges, we introduce AutoGluon-Multimodal (AutoMM)—a Python-based open-source AutoML framework tailored for multimodal learning with foundation models. Embedded within the AutoGluon ecosystem \citep{erickson2020autogluon, shchur2023autogluon}, AutoMM enables users to fine-tune foundation models effortlessly on domain-specific data with just three lines of code. Leveraging popular model repositories such as Huggingface/transformers \citep{wolf2020transformers}, TIMM \citep{rw2019timm}, and MMDetection \citep{chen2019mmdetection}, AutoMM supports a wide range of modalities including text, image, and tabular data, facilitating tasks such as classification, regression, object detection \citep{zou2023object}, semantic matching \citep{reimers2019sentence}, and image segmentation \citep{minaee2021image}. Figure~\ref{fig:automm-intro} outlines the AutoMM framework, illustrating its key functionalities.

The evaluation of AutoMM presents two primary challenges: the absence of established benchmark datasets covering multiple modalities and tasks, and the scarcity of competing AutoML libraries with comparable functionalities. To address the former, we curated a benchmark comprising 55 publicly available datasets spanning diverse modalities and tasks, prioritizing real-world applications over academic datasets for a more robust evaluation of AutoML toolboxes. Mitigating the latter challenge, we conducted a comprehensive evaluation encompassing basic and advanced tasks. In basic classification and regression tasks across 24 unimodal and multimodal datasets, AutoMM outperformed AutoKeras \citep{jin2023autokeras} significantly. For advanced tasks of semantic matching and semantic segmentation, comparisons with task-specific open-source libraries demonstrated comparable performance. These findings underscore the potential of AutoMM as a comprehensive solution for practitioners seeking automated solutions across various modalities and tasks.

\section{Related Work}

AutoML \citep{yao2018taking, zoller2021benchmark, he2021automl} has emerged as a pivotal area of research and development, aiming to democratize machine learning by automating the complex processes of model selection \citep{arango2023quick}, hyperparameter tuning \citep{bischl2023hyperparameter}, and feature engineering \citep{zheng2018feature}. Traditional AutoML methods primarily focus on basic classification and regression tasks for tabular data. For example, Auto-Sklearn \citep{feurer2015efficient} is built on the scikit-learn library, and leverages bayesian optimization, meta-learning and ensemble construction to provide a robust solution for classification and regression tasks. TPOT \citep{olson2016tpot} also uses scikit-learn as its ML library and employs genetic algorithms to optimize entire ML pipelines, including preprocessing and modeling steps. H2O AutoML \citep{ledell2020h2o} provides an end-to-end platform for automating the machine learning process, including automatic training and ensemble of a diverse set of algorithms such as GBMs, Rnadom Forests, Deep Neural Network, GLMs. FLAML \citep{wang2021flaml} optimizes for low computational cost in the hyperparameter search.

Despite the significant strides made in traditional AutoML domains, the exploration of AutoML capabilities for handling multimodal inputs and tackling advanced tasks remains relatively nascent. While several conventional AutoML frameworks \citep{vakhrushev2021lightautoml, feurer2015efficient} are equipped to process multimodal data, their methodologies predominantly hinge on traditional machine learning algorithms. AutoKeras, an AutoML framework leveraging deep learning \citep{jin2023autokeras}, broadens its applicability to multimodal inputs. However, this functionality requires users to manually specify and preprocess for each modality, which increases the likelihood of errors. Additionally, AutoKeras falls short in addressing more complex downstream tasks, such as object detection \citep{zou2023object} and image segmentation \citep{minaee2021image}, limiting its utility in advanced applications. LightAutoML \citep{vakhrushev2021lightautoml} is a lightweight framework which provides fast hyperparameter search and focuses on limited models such as gradient boosted decision trees and linear models for multimodal data. However, its support for multimodal input is restricted to tabular-image and tabular-text only, which limits its usage in real-world applications.

There are also AutoML methods based on foundation models. For example, Quick-Tune \citep{quicktune} proposes how and which models to fine-tune from a pre-trained zoo. TabPFN \citep{hollmann2022tabpfn} proposes to use transformers to solve small tabular classification problems without hyperparameter tuning. OptFormer \citep{optformer} leverages transformers to learn universal hyperparameter optimizers. However, these methods all focused on unimodal data.
In contrast, AutoMM is designed to proficiently manage multimodal inputs, including image, text, and tabular, in a unified framework with three lines of codes. Beyond the basic classification and regression tasks, AutoMM also supports real-world application tasks such as object detection, image segmentation, semantic matching \citep{reimers2019sentence}, etc. At its core, AutoMM leverages the transformative power of recent foundation models \citep{bommasani2021opportunities}, capitalizing on their exceptional transfer learning capabilities to achieve state-of-the-art results across a diverse array of tasks.

\section{AutoMM}
Supporting diverse data modalities and task types, while staying true to the AutoML philosophy, presents significant challenges. The primary obstacle lies in automating these processes through unified data format/processing, APIs, model design, and training workflow. In this section, we delineate the design and functionalities of AutoMM.

\subsection{Data Format and Processing}
AutoMM employs Pandas DataFrame \citep{reback2020pandas} to consolidate various data modalities, including images, text, and tabular (numeric and categorical) data. DataFrames are ubiquitous in modern data analytics due to their flexibility and user-friendly nature. Essentially, a Pandas DataFrame organizes data into a 2-dimensional table, with rows representing individual samples and columns representing features. Each field within the DataFrame can accommodate different data types, such as numeric, categorical, text, image paths, bytearray images, or base64-encoded images. This versatility allows users to provide AutoMM with multimodal DataFrames containing any combination of modalities. Moreover, each modality can comprise multiple columns, facilitating complex data representations. As an AutoML toolbox, AutoMM adeptly handles raw and noisy data, alleviating users from the preprocessing burden.

\begin{figure}[t]
    \centering
    \includegraphics[width=0.9\linewidth,bb=0 0 950 100]{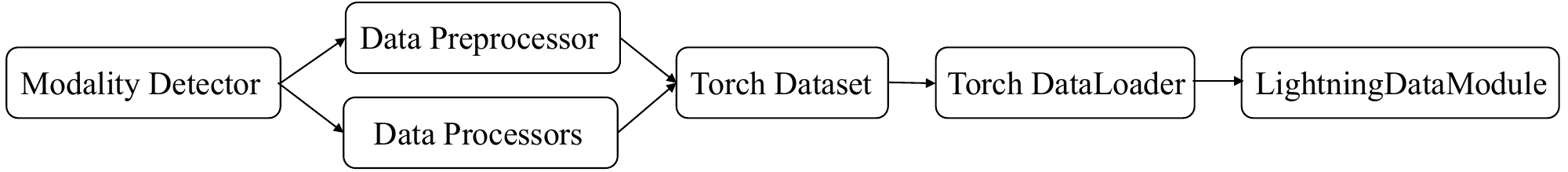}
    \caption{Interdependency among data modules: Each module from left to right serves as a prerequisite for the subsequent one.}
    \label{fig:data-modules}
\end{figure}

Traditionally, addressing different modalities entails constructing independent data processing pipelines for each. However, such an approach leads to code redundancy and increased maintenance overhead. To mitigate this, we systematically analyzed these pipelines and abstracted their commonalities, resulting in a generalized pipeline. Our unified data pipeline comprises LightningDataModule \citep{falcon2019pytorch}, torch \citep{paszke2019pytorch} DataLoader, torch Dataset, a modality detector, a data preprocessor, and arbitrary data processors. At a high level, the LightningDataModule invokes torch DataLoader to generate distinct data loaders for training, validation, and prediction stages. The creation of torch DataLoader necessitates a torch Dataset object, which, in turn, relies on the data preprocessor and data processors. These components are instantiated based on the detected modalities by the modality detector. The interdependence among these data modules is illustrated in Figure~\ref{fig:data-modules}.

The data preprocessor handles model-agnostic data processing tasks, such as filtering non-informative features, converting data types, handling null values, and normalizing numeric data. In contrast, data processors cater to model-specific processing requirements. Since multiple models may operate concurrently, with each potentially accepting multiple modalities as inputs, we create a dedicated data processor for each modality of every model. The data preprocessor conducts DataFrame-level preprocessing during torch Dataset initialization, while data processors perform online transformation of individual samples during data loading. Additionally, we define data collator functions to aggregate processed samples into mini-batches for model consumption.

Data processors are responsible for per-sample processing, encompassing data augmentation \citep{shorten2019survey, shorten2021text} and multi-field processing. Numeric fields are concatenated into a single vector, while categorical fields are encoded separately using torch nn.Embedding. Image processing involves augmentation (e.g., TrivialAugment \citep{muller2021trivialaugment}), tensor conversion, and normalization. In cases where multiple images per sample are present, AutoMM processes each image individually before stacking them. Text data undergoes tokenization, with each text field tokenized separately and concatenated into a unified token sequence. To address sequence length constraints imposed by models, we truncate the sequence by iteratively removing tokens from the longest text field until compliance is achieved.

\subsection{APIs}
\label{sec:api}
AutoMM streamlines the fine-tuning of foundation models on unimodal or multimodal data through a concise API interface. 
To illustrate, let's focus on basic classification/regression tasks using a DataFrame stored in "train.csv" with the label column "label". With just three lines of code, users can import \href{https://auto.gluon.ai/stable/api/autogluon.multimodal.MultiModalPredictor.html}{\color{blue}{\texttt{MultiModalPredictor}}}, initialize an object, and begin training:
\begin{lstlisting}[language=Python]
from autogluon.multimodal import MultiModalPredictor
predictor = MultiModalPredictor(label="label")
predictor.fit("train.csv")
\end{lstlisting}
During training, AutoMM automatically infers the problem type (e.g., binary classification, multi-class classification, or regression), partitions the data into training and validation sets, identifies data modalities, selects appropriate foundation models, and performs fine-tuning. The \href{https://auto.gluon.ai/stable/api/autogluon.multimodal.MultiModalPredictor.fit.html}{\color{blue}{\texttt{fit()}}} method also offers additional customization options, such as controlling training time, customizing hyperparameters , or conducting hyperparameter optimization. Continuous training is supported, enabling sequential calls of \href{https://auto.gluon.ai/stable/api/autogluon.multimodal.MultiModalPredictor.fit.html}{\color{blue}{\texttt{fit()}}} on incoming training data.

% (\texttt{time\_limits}) (\texttt{hyperparameters}) ({hyperparameter\_tune\_kwargs})

Upon completion of training, users can leverage various APIs for evaluation or inference:
\begin{lstlisting}[language=Python]
score = predictor.evaluate("test.csv")
predictions = predictor.predict("new.csv")
probabilities = predictor.predict_proba("new.csv")
embeddings = predictor.extract_embedding("new.csv")
\end{lstlisting}
These APIs facilitate model evaluation, prediction, probability estimation (for classification tasks), and feature embedding extraction. Notably, prediction-related APIs (\href{https://auto.gluon.ai/stable/api/autogluon.multimodal.MultiModalPredictor.predict.html}{\color{blue}{\texttt{predict()}}}, \href{https://auto.gluon.ai/stable/api/autogluon.multimodal.MultiModalPredictor.predict_proba.html}{\color{blue}{\texttt{predict\_proba()}}}, and \href{https://auto.gluon.ai/stable/api/autogluon.multimodal.MultiModalPredictor.extract_embedding.html}{\color{blue}{\texttt{extract\_embedding()}}}) do not necessitate labels, enhancing their utility for deployment purposes. Predictors can be saved and loaded for future use via the \href{https://auto.gluon.ai/stable/api/autogluon.multimodal.MultiModalPredictor.save.html}{\color{blue}{\texttt{save()}}} and \href{https://auto.gluon.ai/stable/api/autogluon.multimodal.MultiModalPredictor.load.html}{\color{blue}{\texttt{load()}}} methods:
\begin{lstlisting}[language=Python]
predictor.save("save_path")
predictor = MultiModalPredictor.load("save_path")
\end{lstlisting}
The \texttt{load()} method also supports training resumption in case of training interruptions by specifying \texttt{resume=True} and providing the path to the interrupted predictor. Calling \href{https://auto.gluon.ai/stable/api/autogluon.multimodal.MultiModalPredictor.fit.html}{\color{blue}{\texttt{fit()}}} afterwards can resume training from the last saved checkpoint. Furthermore, we provide various additional resources on the official website \href{https://auto.gluon.ai}{\color{blue}{auto.gluon.ai}}, including \href{https://auto.gluon.ai/stable/install.html}{\textcolor{blue}{installation instructions}}, \href{https://auto.gluon.ai/stable/tutorials/multimodal/index.html}{\textcolor{blue}{hands-on tutorials}}, and a \href{https://auto.gluon.ai/stable/cheatsheet.html#multimodal}{\textcolor{blue}{cheatsheet}} summarizing the main features.

\begin{figure}[t]
   \begin{minipage}{0.48\textwidth}
     \centering
         \includegraphics[width=.99\linewidth]{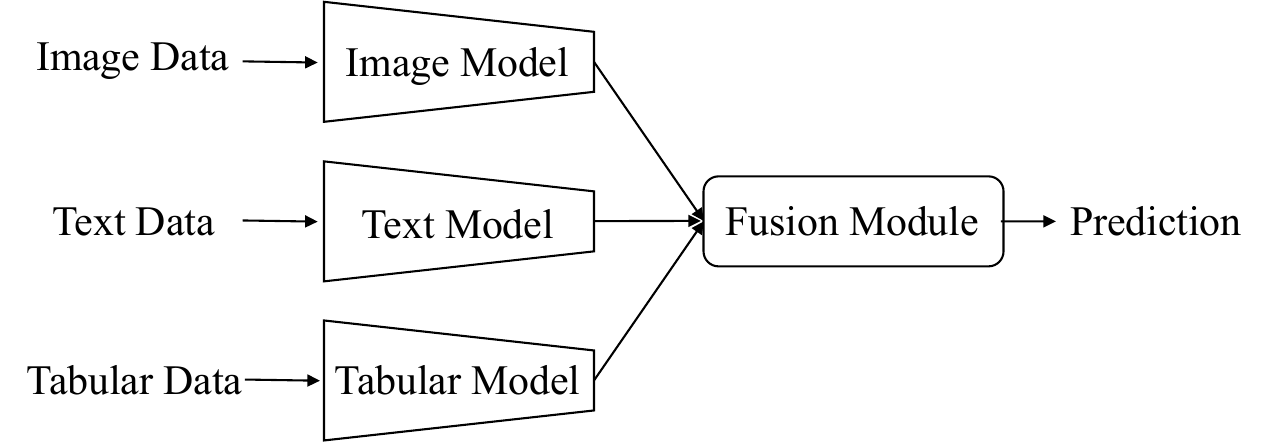}
         \caption{Late-fusion model.}\label{fig:late-fusion-model}
   \end{minipage}\hfill
   \begin{minipage}{0.48\textwidth}
     \centering
     \includegraphics[width=.99\linewidth]{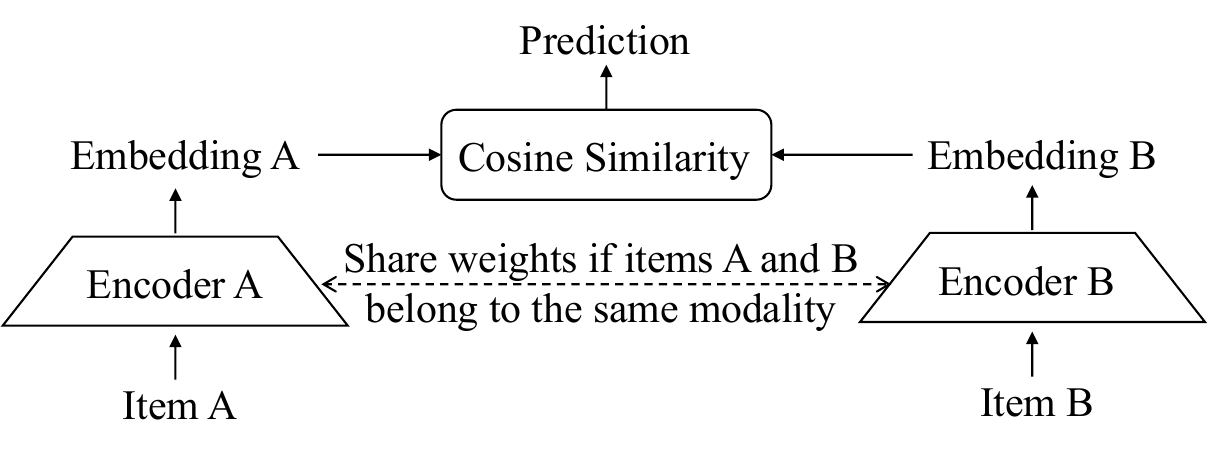}
     \caption{Bi-encoder model.}\label{fig:bi-encoder-model}
   \end{minipage}
\end{figure}

\subsection{Models}
AutoMM aims to apply foundation models across diverse real-world scenarios. Foundation models undergo pretraining on extensive datasets before fine-tuning on smaller downstream labeled datasets. However, this fine-tuning necessitates alignment between downstream data modalities and those encountered during pretraining. For instance, while BERT \citep{devlin2018bert} can fine-tune on text-only tasks, it cannot handle text+tabular data directly. Despite the proliferation of foundation models, their pretraining modalities typically span image-only \citep{dosovitskiy2020image}, text-only \citep{he2021debertav3}, or image+text domains \citep{radford2021learning}. Yet, practical applications frequently involve more combinations such as image+text+tabular. To bridge this gap, AutoMM adopts a late-fusion architecture, scalable to accommodate arbitrary modality combinations. The late-fusion framework, shown in Figure~\ref{fig:late-fusion-model}, integrates independent backbones for images, text, and tabular data, followed by a fusion module (e.g., MLP or transformer \citep{vaswani2017attention} layers) for feature fusion. In scenarios involving a single modality, the fusion module is bypassed. Notably, adding support for a new modality seamlessly enables integration with existing ones.

AutoMM offers extensive model zoo compatibility, including Huggingface/transformers \citep{wolf2020transformers}, TIMM \citep{rw2019timm}, and MMDetection \citep{chen2019mmdetection}. These repositories contain lots of pretrained models with >25000, >700, and >300 respectively, as documented. Text models (e.g., Electra \citep{clark2020electra}, Deberta \citep{he2021debertav3}) from Huggingface and image models (e.g., Swin Transformer \citep{liu2021swin}, ViT \citep{dosovitskiy2020image}) from TIMM are predominantly utilized. MMDetection supplies pretrained models for object detection tasks, such as YOLOX \citep{yolox} and DINO \citep{dino}. Pretrained models generally span a spectrum of sizes (e.g., ViT-large, ViT-base, ViT-small), each associated with distinct performance and resource considerations. AutoMM defines three preset levels (\texttt{best\_quality}, \texttt{high\_quality}, \texttt{medium\_quality}) to accommodate varying performance-latency trade-offs. The selection of preset models is based on our internal benchmarking results, streamlining model selection for end-users. For the basic classification/regression, we employ a two-stage selection process. Initially, we individually select the top 5 models for image, text, and tabular data based on their unimodal benchmarking performance. Subsequently, we conduct a random search of their combinations, along with other hyperparameters like learning rate and epochs, on multimodal benchmarks. The resulting best combination serves as the default models.

\subsection{Training}
By default, AutoMM trains only one late-fusion model with fixed hyperparameters (without optimization). These default hyperparameters, e.g., learning rate and weight decay, have been pre-determined through offline search across diverse benchmark datasets, assuming they are likely to yield reasonable performance on new (similar) datasets. While some AutoML toolboxes like Autosklearn 2.0 \citep{feurer2022auto} also use the pre-determined hyperparameters, our method differs in that it emphasizes offline tuning, rather than further hyperparameter optimization or model ensembling \citep{ erickson2020autogluon}.  Such methods often require training and evaluating numerous models, assuming each trial has relatively low cost. However, fine-tuning foundation models on multimodal data is generally resource-intensive and may not align with this assumption. To address practical concerns, we opt for offline hyperparameter search across our benchmark datasets, enabling end-users to experience efficiency and effectiveness. Detailed hyperparameters are provided in Section \ref{sec:presets} for reference.

AutoMM leverages the Lightning framework \citep{falcon2019pytorch}, built on top of PyTorch \citep{paszke2019pytorch}, to streamline the training workflow. Lightning abstracts key training components into LightningDataModule, LightningModule, Trainer, and Callbacks, enhancing maintainability and scalability. LightningDataModule oversees data loading and processing, employing our data preprocessor and processors. LightningModule defines the model's forward pass and optimizer configuration. Encouraging modular design, our model is created outside LightningModule, being passed into it as needed. Trainer orchestrates the training process, offering extensive configuration options (e.g., gradient accumulation, training precision) and driving training through interaction with LightningDataModule and LightningModule objects. We also utilize Lightning's build-in Callbacks that can provide additional functionalities during the training loop, including logging, checkpointing, and early stopping.

Given the rapid expansion of foundation model sizes, fine-tuning these models poses challenges, particularly for users with limited computational resources. To mitigate this, AutoMM embraces parameter-efficient fine-tuning techniques (PEFT) \citep{houlsby2019parameter}. These techniques, such as BitFit \citep{zaken2021bitfit} and LoRA \citep{hu2021lora}, either optimize a tiny fraction of pretrained weights or introduce lightweight structures on top of fixed pretrained models to reduce memory footprint and training time while preserving performance. For instance, IA3 \citep{liu2022few} facilitates fine-tuning of the Flan-T5-XL \citep{chung2022scaling} model's encoder (1.2 billion parameters) on a single NVIDIA T4 GPU with ~15 GB memory. Additionally, AutoMM supports distributed training (multi-GPU or multi-node) and low-precision (default 16-mixed) training \citep{micikevicius2017mixed}, further easing the burden posed by large models.
% PEFT methods are customizable through the fit() API's hyperparameter settings. 

\subsection{Deployment}
Efficient deployment of trained models into production environments is crucial for real-world applications. To ensure seamless predictor loading in an offline environment, AutoMM pre-saves requisite artifacts—e.g., Huggingface model configuration—forestalling Internet access errors upon predictor invocation. Notably, in production settings, low inference latency, particularly for online inference, is paramount. Although our training hinges on Lightning modules adept at handling large sample sizes, these modules may impede inference when processing few samples. To address this, we furnish a realtime option, eschewing Lightning modules in favor of plain PyTorch models and data processing to expedite inference. Moreover, AutoMM integrates with NVIDIA TensorRT \citep{vanholder2016efficient}, encompassing a deep learning inference optimizer and runtime renowned for low latency and high throughput. Additionally, accommodating varying image formats between training and deployment—e.g., image path in training versus image bytearray in deployment—AutoMM dynamically infers image sub-types during inference, ensuring seamless model deployment.

\subsection{Advanced Tasks}
AutoMM expands traditional AutoML beyond basic classification and regression tasks to encompass advanced functionalities. As of the time of writing, AutoMM supports tasks such as semantic matching, object detection, and semantic segmentation. Semantic matching involves assessing the similarity between two items, which can be two images, two texts, or an image-text pair. We utilize a bi-encoder design, illustrated in Figure~\ref{fig:bi-encoder-model}, that independently encodes the two items in the embedding space before computing their semantic similarity. Object detection and semantic segmentation are both computer vision tasks. Object detection locates object instances within an image, while semantic segmentation categorizes each pixel in an image into a class or object. In terms of output, object detection provides object class labels and bounding boxes, whereas segmentation generates segmentation masks. Despite the complexity of these tasks, users can implement solutions with just three lines of code, albeit with the necessity to specify the \texttt{problem\_type} argument when initializing the predictor object. AutoMM achieves these advanced functionalities by fine-tuning foundation models from Huggingface and MMDetection. For instance, it can employ CLIP \citep{radford2021learning} for image-text matching, DINO \citep{dino} for object detection, and the Segment Anything Model \citep{kirillov2023segment} for semantic segmentation.

\section{Experiments}
\label{sec:exp}

In this experimental study, we aim to assess the efficacy of AutoMM compared to other leading AutoML solutions or task-specific toolboxes across a diverse range of tasks, including classification or regression, semantic matching, object detection, and semantic segmentation. 

\subsection{Classification and Regression}
We start by experimenting how AutoMM performs on classification and regression tasks.  %This is one of the most common applications, as it allows models to leverage diverse sources of information, potentially leading to more accurate predictions and richer insights. the ``default setting'' of
We apply AutoMM to 24 real-life datasets with an assortment of tasks. The input in each task may involve any combinations of \{Tabular, Image, Text\}.  %and our results are shown in Table~\ref{tab:basic_main}. 
The performance metrics we reported are \emph{R-squared} ($R^2$), \emph{F1\_weighted} and \emph{F1} for regression, multi-class and binary classification tasks respectively.

%We considered several AutoML frameworks, e.g. Auto-Sklearn, Auto-Keras, and H2O AutoML.  
To the best of our knowledge, the only other AutoML framework that supports multimodal input is Auto-Keras, serving as the main baseline here. The results, which were run over 5 independent random seeds ($22$, $41$, $54$, $86$, $92$) shown in Table~\ref{tab:basic_main}, indicate that AutoMM works out-of-the-box on all datasets with no prior knowledge or dataset-specific configuration. Compared to Auto-Keras,  AutoMM performs  better in all 24 datasets across different problem types and data modalities, with \emph{statistical significance}.  We also observe that the performance of Auto-Keras often vary drastically over different runs, while AutoMM provides substantially more consistency and reliability. Furthermore, AutoMM excels in ease of use: datasets can simply be loaded into a Pandas DataFrame, and modalities are automatically detected by AutoMM. For image modalities, AutoMM accepts both image paths and image bytearrays, without the need for further processing.  In contrast, Auto-Keras requires that data be reformatted into numpy arrays by the user, and modalities must be explicitly specified, such as \emph{autokeras.ImageInput()}, \emph{autokeras.StructuredDataInput()}, and \emph{autokeras.TextInput()}, which can introduce inconsistency and human errors. For further details on datasets, baseline methods, metrics, and additional setup information, please refer to the appendix.

\begin{table}[tbh]
\centering
\resizebox{1.0\textwidth}{!}{
\begin{tabular}{llllll|ll}
\toprule
                       Dataset & \multicolumn{1}{l}{Text} & \multicolumn{1}{l}{Image} & \multicolumn{1}{l}{Tabular} & Problem Type &Metric& Auto-Keras & \multicolumn{1}{l}{AutoMM}    \\\hline
fashion\_mnist          & \xmark                        & \cmark                         & \xmark                           & Multiclass   & F1\_weighted$\uparrow$                    & 0.876(0.020)                   & \textbf{0.953}(0.002)                  \\
food101                 & \xmark                        & \cmark                         & \xmark                           & Multiclass   & F1\_weighted$\uparrow$                      & 0.024(0.045)                   & \textbf{0.937}(0.001)                  \\
Stanford\_cars          & \xmark                        & \cmark                         & \xmark                           & Multiclass   & F1\_weighted$\uparrow$                    & 0.055(0.079)                   & \textbf{0.892}(0.002)                \\
magnetic\_tile\_defects & \xmark                        & \cmark                         & \xmark                           & Multiclass   & F1\_weighted$\uparrow$                    & 0.627(0.171)                   & \textbf{0.956}(0.014)                   \\
European\_flood\_depth  & \xmark                        & \cmark                         & \xmark                           & Binary     & F1$\uparrow$                              & 0.750(0.017)                   & \textbf{0.790}(0.008)                   \\
Oxford\_flowers         & \xmark                        & \cmark                         & \xmark                           & Multiclass   & F1\_weighted$\uparrow$                    & 0.123(0.155)                   & \textbf{0.989}(0.003)                  \\
OxfordIIITPet           & \xmark                        & \cmark                         & \xmark                           & Multiclass   & F1\_weighted$\uparrow$                    & 0.157(0.283)                   & \textbf{0.958}(0.003)                \\
CD18\_cellphone         & \xmark                        & \cmark                         & \xmark                           & Regression    & R$^2\uparrow$                        & -18.390(35.120)                & \textbf{-1.843}(4.477)           \\
HAM10000                & \xmark                        & \cmark                         & \xmark                           & Multiclass    & F1\_weighted$\uparrow$                    & 0.276(0.211)                   & \textbf{0.608}(0.014)                  \\
hateful\_meme           & \cmark                        & \cmark                         & \xmark                           & Binary     & F1$\uparrow$                              & 0.572(0.099)                   & \textbf{0.596}(0.013)                    \\
petfinder               & \cmark                        & \cmark                         & \cmark                           & Multiclass & F1\_weighted$\uparrow$                    & 0.243 (0.040)                  & \textbf{0.408}(0.006)                  \\
memotion                & \cmark                        & \cmark                         & \cmark                           & Multiclass    & F1\_weighted$\uparrow$                    & 0.297 (0.026)                 & \textbf{0.467}(0.013)                 \\
financial\_news         & \cmark                        & \xmark                         & \xmark                           & Multiclass    & F1\_weighted$\uparrow$                   & 0.678(0.027)                   & \textbf{0.874}(0.010)             \\
MLDoc-11000             & \cmark                        & \xmark                         & \xmark                           & Multiclass    & F1\_weighted$\uparrow$                   & 0.916(0.006)                   & \textbf{0.978}(0.002)             \\
gnad10                  & \cmark                        & \xmark                         & \xmark                           & Multiclass      & F1\_weighted$\uparrow$                   & 0.521(0.029)                   & \textbf{0.899}(0.006)          \\
MultiATIS-5000          & \cmark                        & \xmark                         & \xmark                           & Multiclass    & F1\_weighted$\uparrow$                   & 0.864(0.010)                   & \textbf{0.990}(0.003)              \\
fb\_dialog              & \cmark                        & \xmark                         & \xmark                           & Multiclass    & F1\_weighted$\uparrow$                   & 0.982(0.003)                   & \textbf{0.992}(0.001)              \\
SNIPS                   & \cmark                        & \xmark                         & \xmark                           & Multiclass     & F1\_weighted$\uparrow$                   & 0.049(0.018)                   & \textbf{0.990}(0.002)             \\
ag\_news                & \cmark                        & \xmark                         & \xmark                           & Multiclass    & F1\_weighted$\uparrow$                   & 0.887(0.004)                   & \textbf{0.944}(0.001)                \\
airbnb\_melbourn        & \cmark                        & \xmark                         & \cmark                           & Multiclass    & F1\_weighted$\uparrow$                   & 0.198(0.071)                   & \textbf{0.397}(0.011)                 \\
kick\_start\_funding    & \cmark                        & \xmark                         & \cmark                           & Binary       & F1$\uparrow$                             & 0.401 (0.151)                  & \textbf{0.609}(0.005)             \\
cloth\_review           & \cmark                        & \xmark                         & \cmark                           & Regression    & R$^2\uparrow$                              & 0.542(0.053)                   & \textbf{0.735}(0.004)             \\
news\_popularity        & \cmark                        & \xmark                         & \cmark                           & Regression       & R$^2\uparrow$                               & -1.306(1.863)                   & \textbf{0.014}(0.003)              \\
California\_house       & \cmark                        & \xmark                         & \cmark                           & Regression    & R$^2\uparrow$                            & -53757156.425 (55682587.109)    & \textbf{0.944}(0.001)            \\
\bottomrule
\end{tabular}
}

\caption{Results for unimodal/multimodal classification and regression.  The mean performance metrics and error bars (in parentheses) with 0.95 coverage are reported for both AutoMM and Auto-Keras. These numbers are estimated using $5$ independent repeats with different random seeds using $1.96 \times \mathrm{std}/\sqrt{\#\text{ of repeats}}$. Boldface indicates results that are better than the competing framework with \emph{statistical significance} at the level 0.05 using Wald test \citep{wald1943tests}.}
\label{tab:basic_main}
\end{table}

\subsection{Semantic Matching}

\begin{figure}
    \centering
    \includegraphics[width=\linewidth]{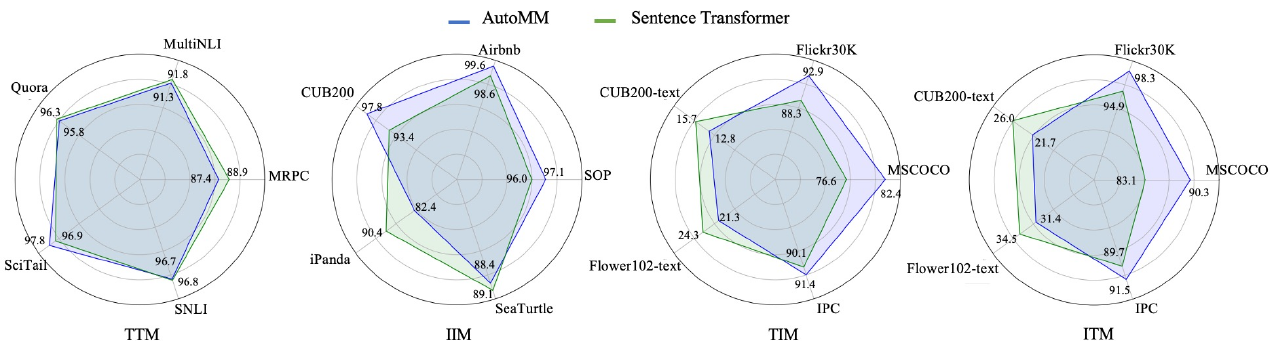}
    \caption{Comparison of AutoMM and Sentence-Transformer on Text to Text Matching (TTM), Image to Image Matching (IIM), Text to Image Matching (TIM), and Image to Text Matching (ITM).}
    \label{fig:semantic_matching}
\end{figure}

For semantic matching, we compare with Sentence-Transformer \citep{reimers2019sentence}, which is a toolbox designed specifically for semantic matching tasks. We show the comparisons on Text to Text Matching (TTM), Image to Image Matching (IIM), Text to Image Matching (TIM), and Image to Text Matching (ITM) tasks in Figure~\ref{fig:semantic_matching}. AutoMM achieves competitive performance with Sentence Transformer, while being more user-friendly and straightforward to utilize. For Sentence Transformer, users have to manually select the optimal loss function for different tasks to achieve the best performance, which requires the user to have extensive domain expertise and a large amount of trial and error. However, AutoMM automates the entire pipeline while being competitive in different tasks. For further details on datasets, baseline methods, metrics, and raw results, please refer to the appendix.

\subsection{Object Detection}

In Table~\ref{tab:det_main}, we select several downstream tasks from various domains and compare AutoMM with baseline frameworks from both Vertex AI\footnote{Vertex AI, Google Cloud, \url{https://cloud.google.com/vertex-ai}.} and NVIDIA TAO\footnote{NVIDIA TAO Toolkit, NVIDIA Developer, \url{https://developer.nvidia.com/tao-toolkit}.}. \textbf{Throughout the comparison, we demonstrate that AutoMM surpasses other AutoML object detection solutions in terms of performance and speed while offering greater ease of use}. While Vertex AI provides efficient model development and deployment tools, it may have limitations in pricing flexibility and customization options compared to other solutions. In our experiments, due to the high cost of Vertex AI, we focused on evaluating it using the "Higher accuracy (new)" option on only four datasets. In contrast, we conducted a thorough comparison with NVIDIA TAO using a pretrained FAN-L-Hybrid backbone~\citep{fan}. It is important to note that both Vertex AI and NVIDIA TAO have specific data requirements, necessitating additional preprocessing beyond the common COCO~\citep{mscoco} format. Furthermore, NVIDIA TAO requires configuration for each dataset. In contrast, AutoMM can be used with just a few lines of code without additional data processing or configuration. For further details on the datasets, baseline methods, metrics, and additional setup information, please refer to the appendix.

\begin{table}[]
\scalebox{0.7}{
\begin{tabular}{lc|ccc|ccc|ccc}
\hline
                              &            & \multicolumn{3}{c|}{Vertex AI}                                             & \multicolumn{3}{c|}{Nvidia Tao}                                                               & \multicolumn{3}{c}{AutoMM}                                                                                                    \\
                              &            & \begin{tabular}[c]{@{}c@{}}training\\ time (hrs)\end{tabular} & mAP & AP50 & \begin{tabular}[c]{@{}c@{}}training\\ time (hrs)\end{tabular} & mAP           & AP50          & \begin{tabular}[c]{@{}c@{}}training\\ time (hrs)\end{tabular} & mAP                                        & AP50                \\ \hline
                              & plantdoc   & 2.8                                                           & /   & 49.1 & 3.5                                                           & 55.5          & 70.8          & \textbf{2.1(0.23)}                                            & \textbf{58.7(0.38)}                        & \textbf{73.3(0.54)} \\
\multirow{-2}{*}{Agriculture} & deepfruits & /                                                             & /   & /    & \textbf{0.6}                                                  & 51.4          & 75.2          & \textbf{0.6(0.07)}                                            & {\color[HTML]{000000} \textbf{71.3(0.56)}} & \textbf{95.0(0.19)} \\ \hline
                              & chest10    & 3.8                                                           & /   & 36.8 & 2.5                                                           & \textbf{19.5} & \textbf{41.4} & \textbf{2.1(0.11)}                                            & 18.3(0.28)                                 & 37.2(1.2)           \\
\multirow{-2}{*}{Medical}     & deeplesion & /                                                             & /   & /    & 23.5                                                          & 39.4          & 67.5          & \textbf{18.7(0.40)}                                           & \textbf{40.4(0.24)}                        & \textbf{70.5(0.49)} \\ \hline
Domain Transfer               & comic      & 2.8                                                           & /   & 57.6 & 1.1                                                           & 22.7          & 42.6          & \textbf{0.9(0.09)}                                            & \textbf{42.3(0.72)}                        & \textbf{67.7(0.62)} \\ \hline
Remote Sencing                & dota       & 11.8                                                          & /   & 60.7 & 10.0                                                          & 47.9          & 74.1          & \textbf{9.0(0.75)}                                            & \textbf{51.3(0.14)}                        & \textbf{76.9(0.18)} \\ \hline
Autonomous Driving            & kitti      & /                                                             & /   & /    & 7.9                                                           & \textbf{78.7} & \textbf{96.4} & \textbf{5.9(0.44)}                                            & 71.5(0.76)                                 & 95.4(0.14)          \\ \hline
Infrared                      & thermal    & /                                                             & /   & /    & \textbf{0.3}                                                  & 66.8          & 82.2          & \textbf{0.3(0.05)}                                            & \textbf{82.9(0.88)}                        & \textbf{95.3(1.08)} \\ \hline
\end{tabular}
}
\caption{Comparison of AutoMM and baseline methods on object detection tasks in various domains. The mean performance metrics and error bars (in parentheses) with 0.95 coverage are reported for AutoMM. These numbers are estimated using $3$ independent repeats with different random seeds using $1.96 \times \mathrm{std}/\sqrt{\#\text{ of repeats}}$.}
\label{tab:det_main}
\end{table}

\subsection{Semantic Segmentation}

\iffalse
Semantic segmentation~\citep{mmseg, detectron2, kirillov2023segment, convlora}, on the other hand, is a pixel-level image analysis task that involves partitioning an image into semantically meaningful regions and assigning a class label to each pixel. Unlike object detection, semantic segmentation provides detailed information about the spatial extent of objects in an image, allowing for fine-grained understanding of the scene's content. This task finds applications in medical image analysis, autonomous navigation, and scene understanding in robotics.

In our study, we conducted a comparative analysis of our AutoML semantic segmentation solution with SAM-Adapter and Detectron2, despite the latter two not being AutoML solutions. Our evaluation aimed to assess the performance of these frameworks on several downstream datasets. Remarkably, our AutoML semantic segmentation solution demonstrated superior performance across multiple metrics, outperforming SAM-Adapter and Detectron2. This highlights the efficacy and competitiveness of our AutoML approach. Additionally, our findings underscore the importance of AutoML in simplifying and enhancing the process of semantic segmentation for various real-world applications.
\fi

In Table~\ref{tab:seg_main}, we compare AutoMM with other open-source semantic segmentation toolboxes across datasets from diverse domains. AutoMM demonstrates superior or comparable performance \textbf{with minimal trainable model parameters}. This is achieved through SAM's parameter-efficient fine-tuning, enabling effective segmentation results within a low parameter budget. Additionally, compared to these toolboxes, AutoMM offers a more streamlined approach to training setup as outlined in Section~\ref{sec:api}, eliminating the need of rewriting dataloaders or configuration files for new datasets. For further details on datasets, baseline methods, metrics, and additional setup information, please refer to the appendix.

\begin{table}[t]\large
    \centering
    \setlength{\tabcolsep}{1.5pt}
    \renewcommand{\arraystretch}{1.5}
    \resizebox{\textwidth}{!}{%
    \begin{tabular}{l|c|cc|cc|c|cc|c|c|c|c}
        \toprule
        \multirow{3}*{\bf \large Method} & \multirow{3}*{\bf \#Params (M)} & \multicolumn{5}{c|}{\bf \large Medical} & \multicolumn{4}{c|}{\bf \large Natural Images} & \multicolumn{1}{c|}{\bf \large Agriculture} & \multicolumn{1}{c}{\bf \large Remote Sensing}\\
        \cline{3-13}
        &  &\multicolumn{2}{c|}{ Kvasir} &\multicolumn{2}{c|}{CVC-612} & \multicolumn{1}{c|}{ISIC 2017} & \multicolumn{2}{c|}{ CAMO} &   SBU & Trans10K-v2 & \multicolumn{1}{c|}{Leaf} & \multicolumn{1}{c}{ Road} \\
        ~ & & $ S_\alpha \uparrow$ & $E_\phi \uparrow$   
        & $ S_\alpha \uparrow$ & $E_\phi \uparrow$ & Jac $\uparrow$ & $ S_\alpha \uparrow$ & $E_\phi \uparrow$ & BER  $\downarrow$& mIoU $\uparrow$  & IoU $\uparrow$ &  IoU $\uparrow$  \\
        \midrule
        Detectron2  & 47.56  &90.4 & 94.5 & 89.6&91.8&76.1&73.4&81.7&7.11& \textbf{70.8} &  66.6 & 54.9 \\
        OpenSeg  & 74.50  & \textbf{92.2} & \textbf{95.4} & \textbf{93.3}&\textbf{95.3}&\textbf{78.2}&76.3&81.1& 7.92 & 66.1  & \textbf{78.9}&  35.3 \\
        % MMSeg  & 215.45  & 92.0 & 95.9 & 92.7 & 97.2 & 80.5 & 88.7 & 94.6 & 4.95 &71.5&  80.6 & 53.4 \\
              AutoMM & 8.80  & {92.1} & {94.7} & {90.5} & {92.5} & {77.9} &  \textbf{89.3} & \textbf{92.9} & \textbf{3.90} & 69.2 & {72.9} & \textbf{62.0} \\

        \bottomrule
    \end{tabular}}

    \caption{Comparison of AutoMM and baselines on semantic segmentation tasks in various domains.}
    \label{tab:seg_main}
\end{table}

\section{Conclusion and Future Work}\label{sec:conclusion-future-work}
This paper introduces AutoMM, an AutoML toolbox with a focus on foundation models and multimodal learning. Key of AutoMM lies in its support for multiple modalities, tasks, and model zoos, achieved through a unified internal pipeline and user-friendly APIs. To evaluate AutoMM’s unique capacities, we also build comprehensive benchmarks covering unimodal and multimodal data, as well as basic and advanced tasks. Experimental results demonstrate AutoMM’s superior performance and ease of use. While AutoMM represents an initial step towards bridging practical AutoML development with cutting-edge AI research, closing the remaining gap requires significant effort. AutoMM is in active development, and several potential directions are being considered:

\textbf{Multimodal Foundation Models}. The current late-fusion model can accommodate various modality combinations, but its performance may be suboptimal because unimodal foundation models are pretrained independently. Using multimodal foundation models that capture modality interactions during pretraining could enhance performance on some downstream tasks.

\textbf{Generative Tasks}. While AutoMM supports discriminative tasks, e.g., classification and regression, there is a need to include generative tasks like image generation and question answering. Given the growing interest in generative AI and the availability of open-source generative foundation models, incorporating support for generative tasks could expand AutoML capabilities significantly.

\textbf{More Modalities}. Many real-world applications involve data modalities beyond images, text, and tabular data. AI research has seen rapid progress in modalities such as documents, audio, and video. Expanding AutoMM to support these modalities and integrating relevant research outcomes can further broaden the scope of AutoML.

\section{Broader Impact Statement}\label{sec:broader-impact}

AutoMM presents a notable advancement in AutoML by providing a unified framework capable of processing multimodal data for complex tasks with minimal input required. It democratizes machine learning, enabling non-experts to utilize advanced capabilities for tasks like object detection and image segmentation. This can spur innovation and enhance productivity across multiple domains.

The potential negative impacts of the proposed approach are similar to those of other methods reliant on foundation models, encompassing issues such as data privacy, security concerns, and the perpetuation of bias in machine learning models. Like these methods, AutoMM's effectiveness and ethical implications are tightly coupled with the characteristics of the underlying data and the design of the foundation models it employs. It is recommended that users must scrutinize their data to prevent the reinforcement of existing biases through the models trained on them.

% The 9 pages allocated for the main paper must include a broader impact
% statement regarding the approach, datasets and applications proposed/used in
% your paper. It should reflect on the environmental, ethical and societal
% implications of your work. The statement should require at most one page and
% must be included both at submission and camera-ready time.
%
% If authors have reflected on their work and determined that there are no
% likely negative broader impacts, they may use the following statement:
%
% After careful reflection, the authors have determined that this work presents
% no notable negative impacts to society or the environment.
%
% This section is included in the template as a default, but you can also place these
% discussions anywhere else in the main paper, e.g., in the introduction/future work.
%
% The Centre for the Governance of AI has written an excellent guide for writing
% good broader impact statements (for the NeurIPS conference) that may be a
% useful resource for AutoML-Conf authors:
%
% https://medium.com/@GovAI/a-guide-to-writing-the-neurips-impact-statement-4293b723f832

\begin{acknowledgements}
  % content in acknowledgements will be automatically hidden during submission
\end{acknowledgements}

% ==== Bibliography
% print bibliography -- for bibtex / natbib, use:

\clearpage
\bibliography{reference}

\begin{thebibliography}{121}
\providecommand{\natexlab}[1]{#1}
\providecommand{\url}[1]{\texttt{#1}}
\expandafter\ifx\csname urlstyle\endcsname\relax
  \providecommand{\doi}[1]{doi: #1}\else
  \providecommand{\doi}{doi: \begingroup \urlstyle{rm}\Url}\fi

\bibitem[air()]{airbnb_melbourne}
{Melbourne Airbnb Open Data}.
\newblock \url{https://www.kaggle.com/datasets/tylerx/melbourne-airbnb-open-data}.
\newblock Accessed: 2024-02-02.

\bibitem[pet()]{petfinder}
{PetFinder.my Adoption Prediction}.
\newblock \url{https://www.kaggle.com/c/petfinder-adoption-prediction/data/}.
\newblock Accessed: 2024-02-02.

\bibitem[Agarap(2018)]{cloth_review}
Abien~Fred Agarap.
\newblock Statistical analysis on e-commerce reviews, with sentiment classification using bidirectional recurrent neural network (rnn).
\newblock \emph{arXiv preprint arXiv:1805.03687}, 2018.

\bibitem[Airbnb(2020)]{airbnb}
Airbnb.
\newblock Airbnb duplicate image dataset, 2020.
\newblock https://www.kaggle.com/datasets/barelydedicated/airbnb-duplicate-image-detection/data.

\bibitem[Arango et~al.(2023{\natexlab{a}})Arango, Ferreira, Kadra, and Grabocka]{arango2023quick}
Sebastian~Pineda Arango, Fabio Ferreira, Arlind Kadra, and Frank Hutter~Josif Grabocka.
\newblock Quick-tune: Quickly learning which pretrained model to finetune and how.
\newblock \emph{arXiv preprint arXiv:2306.03828}, 2023{\natexlab{a}}.

\bibitem[Arango et~al.(2023{\natexlab{b}})Arango, Ferreira, Kadra, and Grabocka]{quicktune}
Sebastian~Pineda Arango, Fabio Ferreira, Arlind Kadra, and Frank Hutter~Josif Grabocka.
\newblock Quick-tune: Quickly learning which pretrained model to finetune and how.
\newblock \emph{arXiv preprint arXiv:2306.03828}, 2023{\natexlab{b}}.

\bibitem[Baltru{\v{s}}aitis et~al.(2018)Baltru{\v{s}}aitis, Ahuja, and Morency]{baltruvsaitis2018multimodal}
Tadas Baltru{\v{s}}aitis, Chaitanya Ahuja, and Louis-Philippe Morency.
\newblock Multimodal machine learning: A survey and taxonomy.
\newblock \emph{IEEE transactions on pattern analysis and machine intelligence}, 41\penalty0 (2):\penalty0 423--443, 2018.

\bibitem[Bernal et~al.(2015)Bernal, S{\'a}nchez, Fern{\'a}ndez-Esparrach, Gil, Rodr{\'\i}guez, and Vilari{\~n}o]{bernal2015wm}
Jorge Bernal, F~Javier S{\'a}nchez, Gloria Fern{\'a}ndez-Esparrach, Debora Gil, Cristina Rodr{\'\i}guez, and Fernando Vilari{\~n}o.
\newblock Wm-dova maps for accurate polyp highlighting in colonoscopy: Validation vs. saliency maps from physicians.
\newblock \emph{Computerized medical imaging and graphics}, 43:\penalty0 99--111, 2015.

\bibitem[Bischl et~al.(2023)Bischl, Binder, Lang, Pielok, Richter, Coors, Thomas, Ullmann, Becker, Boulesteix, et~al.]{bischl2023hyperparameter}
Bernd Bischl, Martin Binder, Michel Lang, Tobias Pielok, Jakob Richter, Stefan Coors, Janek Thomas, Theresa Ullmann, Marc Becker, Anne-Laure Boulesteix, et~al.
\newblock Hyperparameter optimization: Foundations, algorithms, best practices, and open challenges.
\newblock \emph{Wiley Interdisciplinary Reviews: Data Mining and Knowledge Discovery}, 13\penalty0 (2):\penalty0 e1484, 2023.

\bibitem[Bommasani et~al.(2021)Bommasani, Hudson, Adeli, Altman, Arora, von Arx, Bernstein, Bohg, Bosselut, Brunskill, et~al.]{bommasani2021opportunities}
Rishi Bommasani, Drew~A Hudson, Ehsan Adeli, Russ Altman, Simran Arora, Sydney von Arx, Michael~S Bernstein, Jeannette Bohg, Antoine Bosselut, Emma Brunskill, et~al.
\newblock On the opportunities and risks of foundation models.
\newblock \emph{arXiv preprint arXiv:2108.07258}, 2021.

\bibitem[Bossard et~al.(2014)Bossard, Guillaumin, and Van~Gool]{food101}
Lukas Bossard, Matthieu Guillaumin, and Luc Van~Gool.
\newblock Food-101 -- mining discriminative components with random forests.
\newblock In \emph{European Conference on Computer Vision}, 2014.

\bibitem[Chan et~al.(2020)Chan, Samala, Hadjiiski, and Zhou]{chan2020deep}
Heang-Ping Chan, Ravi~K Samala, Lubomir~M Hadjiiski, and Chuan Zhou.
\newblock Deep learning in medical image analysis.
\newblock \emph{Deep learning in medical image analysis: challenges and applications}, pages 3--21, 2020.

\bibitem[Chen et~al.(2019)Chen, Wang, Pang, Cao, Xiong, Li, Sun, Feng, Liu, Xu, et~al.]{chen2019mmdetection}
Kai Chen, Jiaqi Wang, Jiangmiao Pang, Yuhang Cao, Yu~Xiong, Xiaoxiao Li, Shuyang Sun, Wansen Feng, Ziwei Liu, Jiarui Xu, et~al.
\newblock Mmdetection: Open mmlab detection toolbox and benchmark.
\newblock \emph{arXiv preprint arXiv:1906.07155}, 2019.

\bibitem[Chen et~al.(2022)Chen, Song, Lee, Wang, Zhang, Dohan, Kawakami, Kochanski, Doucet, Ranzato, et~al.]{optformer}
Yutian Chen, Xingyou Song, Chansoo Lee, Zi~Wang, Richard Zhang, David Dohan, Kazuya Kawakami, Greg Kochanski, Arnaud Doucet, Marc'aurelio Ranzato, et~al.
\newblock Towards learning universal hyperparameter optimizers with transformers.
\newblock \emph{Advances in Neural Information Processing Systems}, 35:\penalty0 32053--32068, 2022.

\bibitem[Chung et~al.(2022)Chung, Hou, Longpre, Zoph, Tay, Fedus, Li, Wang, Dehghani, Brahma, et~al.]{chung2022scaling}
Hyung~Won Chung, Le~Hou, Shayne Longpre, Barret Zoph, Yi~Tay, William Fedus, Yunxuan Li, Xuezhi Wang, Mostafa Dehghani, Siddhartha Brahma, et~al.
\newblock Scaling instruction-finetuned language models.
\newblock \emph{arXiv preprint arXiv:2210.11416}, 2022.

\bibitem[Ciaglia et~al.(2022)Ciaglia, Zuppichini, Guerrie, McQuade, and Solawetz]{roboflow100}
Floriana Ciaglia, Francesco~Saverio Zuppichini, Paul Guerrie, Mark McQuade, and Jacob Solawetz.
\newblock Roboflow 100: A rich, multi-domain object detection benchmark.
\newblock \emph{arXiv preprint arXiv:2211.13523}, 2022.

\bibitem[Clark et~al.(2020)Clark, Luong, Le, and Manning]{clark2020electra}
Kevin Clark, Minh-Thang Luong, Quoc~V Le, and Christopher~D Manning.
\newblock Electra: Pre-training text encoders as discriminators rather than generators.
\newblock \emph{arXiv preprint arXiv:2003.10555}, 2020.

\bibitem[Codella et~al.(2018)Codella, Gutman, Celebi, Helba, Marchetti, Dusza, Kalloo, Liopyris, Mishra, Kittler, et~al.]{codella2018skin}
Noel~CF Codella, David Gutman, M~Emre Celebi, Brian Helba, Michael~A Marchetti, Stephen~W Dusza, Aadi Kalloo, Konstantinos Liopyris, Nabin Mishra, Harald Kittler, et~al.
\newblock Skin lesion analysis toward melanoma detection: A challenge at the 2017 international symposium on biomedical imaging (isbi), hosted by the international skin imaging collaboration (isic).
\newblock In \emph{2018 IEEE 15th international symposium on biomedical imaging (ISBI 2018)}, pages 168--172. IEEE, 2018.

\bibitem[Contributors(2019)]{Openseg}
Openseg Contributors.
\newblock Openseg: The official pytorch implementation of ocnet series and segfix.
\newblock \url{https://github.com/openseg-group/openseg.pytorch/tree/pytorch-1.7}, 2019.

\bibitem[Cordts et~al.(2016)Cordts, Omran, Ramos, Rehfeld, Enzweiler, Benenson, Franke, Roth, and Schiele]{cordts2016cityscapes}
Marius Cordts, Mohamed Omran, Sebastian Ramos, Timo Rehfeld, Markus Enzweiler, Rodrigo Benenson, Uwe Franke, Stefan Roth, and Bernt Schiele.
\newblock The cityscapes dataset for semantic urban scene understanding.
\newblock In \emph{Proceedings of the IEEE conference on computer vision and pattern recognition}, pages 3213--3223, 2016.

\bibitem[Coucke et~al.(2018)Coucke, Saade, Ball, Bluche, Caulier, Leroy, Doumouro, Gisselbrecht, Caltagirone, Lavril, et~al.]{snips}
Alice Coucke, Alaa Saade, Adrien Ball, Th{\'e}odore Bluche, Alexandre Caulier, David Leroy, Cl{\'e}ment Doumouro, Thibault Gisselbrecht, Francesco Caltagirone, Thibaut Lavril, et~al.
\newblock Snips voice platform: an embedded spoken language understanding system for private-by-design voice interfaces.
\newblock \emph{arXiv preprint arXiv:1805.10190}, 2018.

\bibitem[Devlin et~al.(2018)Devlin, Chang, Lee, and Toutanova]{devlin2018bert}
Jacob Devlin, Ming-Wei Chang, Kenton Lee, and Kristina Toutanova.
\newblock Bert: Pre-training of deep bidirectional transformers for language understanding.
\newblock \emph{arXiv preprint arXiv:1810.04805}, 2018.

\bibitem[Dolan and Brockett(2005)]{mrpc}
Bill Dolan and Chris Brockett.
\newblock Automatically constructing a corpus of sentential paraphrases.
\newblock In \emph{Third International Workshop on Paraphrasing (IWP2005)}, 2005.

\bibitem[Dosovitskiy et~al.(2020)Dosovitskiy, Beyer, Kolesnikov, Weissenborn, Zhai, Unterthiner, Dehghani, Minderer, Heigold, Gelly, et~al.]{dosovitskiy2020image}
Alexey Dosovitskiy, Lucas Beyer, Alexander Kolesnikov, Dirk Weissenborn, Xiaohua Zhai, Thomas Unterthiner, Mostafa Dehghani, Matthias Minderer, Georg Heigold, Sylvain Gelly, et~al.
\newblock An image is worth 16x16 words: Transformers for image recognition at scale.
\newblock \emph{arXiv preprint arXiv:2010.11929}, 2020.

\bibitem[Erickson et~al.(2020)Erickson, Mueller, Shirkov, Zhang, Larroy, Li, and Smola]{erickson2020autogluon}
Nick Erickson, Jonas Mueller, Alexander Shirkov, Hang Zhang, Pedro Larroy, Mu~Li, and Alexander Smola.
\newblock Autogluon-tabular: Robust and accurate automl for structured data.
\newblock \emph{arXiv preprint arXiv:2003.06505}, 2020.

\bibitem[Falcon(2019)]{falcon2019pytorch}
William~A Falcon.
\newblock Pytorch lightning.
\newblock \emph{GitHub}, 3, 2019.

\bibitem[Fan et~al.(2017)Fan, Cheng, Liu, Li, and Borji]{fan2017structure}
Deng-Ping Fan, Ming-Ming Cheng, Yun Liu, Tao Li, and Ali Borji.
\newblock Structure-measure: A new way to evaluate foreground maps.
\newblock In \emph{Proceedings of the IEEE international conference on computer vision}, pages 4548--4557, 2017.

\bibitem[Fan et~al.(2018)Fan, Gong, Cao, Ren, Cheng, and Borji]{fan2018enhanced}
Deng-Ping Fan, Cheng Gong, Yang Cao, Bo~Ren, Ming-Ming Cheng, and Ali Borji.
\newblock Enhanced-alignment measure for binary foreground map evaluation.
\newblock \emph{arXiv preprint arXiv:1805.10421}, 2018.

\bibitem[Fan et~al.(2020)Fan, Ji, Sun, Cheng, Shen, and Shao]{fan2020camouflaged}
Deng-Ping Fan, Ge-Peng Ji, Guolei Sun, Ming-Ming Cheng, Jianbing Shen, and Ling Shao.
\newblock Camouflaged object detection.
\newblock In \emph{Proceedings of the IEEE/CVF conference on computer vision and pattern recognition}, pages 2777--2787, 2020.

\bibitem[Fang et~al.(2024)Fang, Han, Zhang, Zhou, Hu, and Ye]{aug}
Haoyang Fang, Boran Han, Shuai Zhang, Su~Zhou, Cuixiong Hu, and Wen-Ming Ye.
\newblock Data augmentation for object detection via controllable diffusion models.
\newblock In \emph{Proceedings of the IEEE/CVF Winter Conference on Applications of Computer Vision}, pages 1257--1266, 2024.

\bibitem[Fernandes et~al.(2015)Fernandes, Vinagre, Cortez, and Sernadela]{news_popularity}
Kelwin Fernandes, Pedro Vinagre, Paulo Cortez, and Pedro Sernadela.
\newblock {Online News Popularity}.
\newblock UCI Machine Learning Repository, 2015.
\newblock {DOI}: https://doi.org/10.24432/C5NS3V.

\bibitem[Feurer et~al.(2015)Feurer, Klein, Eggensperger, Springenberg, Blum, and Hutter]{feurer2015efficient}
Matthias Feurer, Aaron Klein, Katharina Eggensperger, Jost Springenberg, Manuel Blum, and Frank Hutter.
\newblock Efficient and robust automated machine learning.
\newblock \emph{Advances in neural information processing systems}, 28, 2015.

\bibitem[Feurer et~al.(2022)Feurer, Eggensperger, Falkner, Lindauer, and Hutter]{feurer2022auto}
Matthias Feurer, Katharina Eggensperger, Stefan Falkner, Marius Lindauer, and Frank Hutter.
\newblock Auto-sklearn 2.0: Hands-free automl via meta-learning.
\newblock \emph{Journal of Machine Learning Research}, 23\penalty0 (261):\penalty0 1--61, 2022.

\bibitem[Gada et~al.(2021)Gada, Haria, Mankad, Damania, and Sankhe]{gada2021automated}
Mihir Gada, Zenil Haria, Arnav Mankad, Kaustubh Damania, and Smita Sankhe.
\newblock Automated feature engineering and hyperparameter optimization for machine learning.
\newblock In \emph{2021 7th International Conference on Advanced Computing and Communication Systems (ICACCS)}, volume~1, pages 981--986. IEEE, 2021.

\bibitem[Ge et~al.(2021)Ge, Liu, Wang, Li, and Sun]{yolox}
Zheng Ge, Songtao Liu, Feng Wang, Zeming Li, and Jian Sun.
\newblock Yolox: Exceeding yolo series in 2021.
\newblock \emph{arXiv preprint arXiv:2107.08430}, 2021.

\bibitem[Geiger et~al.(2012)Geiger, Lenz, and Urtasun]{kitti}
Andreas Geiger, Philip Lenz, and Raquel Urtasun.
\newblock Are we ready for autonomous driving? the kitti vision benchmark suite.
\newblock In \emph{Conference on Computer Vision and Pattern Recognition (CVPR)}, 2012.

\bibitem[He et~al.(2021{\natexlab{a}})He, Gao, and Chen]{he2021debertav3}
Pengcheng He, Jianfeng Gao, and Weizhu Chen.
\newblock Debertav3: Improving deberta using electra-style pre-training with gradient-disentangled embedding sharing.
\newblock \emph{arXiv preprint arXiv:2111.09543}, 2021{\natexlab{a}}.

\bibitem[He et~al.(2021{\natexlab{b}})He, Zhao, and Chu]{he2021automl}
Xin He, Kaiyong Zhao, and Xiaowen Chu.
\newblock Automl: A survey of the state-of-the-art.
\newblock \emph{Knowledge-based systems}, 212:\penalty0 106622, 2021{\natexlab{b}}.

\bibitem[Hollmann et~al.(2022)Hollmann, M{\"u}ller, Eggensperger, and Hutter]{hollmann2022tabpfn}
Noah Hollmann, Samuel M{\"u}ller, Katharina Eggensperger, and Frank Hutter.
\newblock Tabpfn: A transformer that solves small tabular classification problems in a second.
\newblock \emph{arXiv preprint arXiv:2207.01848}, 2022.

\bibitem[Houlsby et~al.(2019)Houlsby, Giurgiu, Jastrzebski, Morrone, De~Laroussilhe, Gesmundo, Attariyan, and Gelly]{houlsby2019parameter}
Neil Houlsby, Andrei Giurgiu, Stanislaw Jastrzebski, Bruna Morrone, Quentin De~Laroussilhe, Andrea Gesmundo, Mona Attariyan, and Sylvain Gelly.
\newblock Parameter-efficient transfer learning for nlp.
\newblock In \emph{International Conference on Machine Learning}, pages 2790--2799. PMLR, 2019.

\bibitem[Hu et~al.(2021)Hu, Shen, Wallis, Allen-Zhu, Li, Wang, Wang, and Chen]{hu2021lora}
Edward~J Hu, Yelong Shen, Phillip Wallis, Zeyuan Allen-Zhu, Yuanzhi Li, Shean Wang, Lu~Wang, and Weizhu Chen.
\newblock Lora: Low-rank adaptation of large language models.
\newblock \emph{arXiv preprint arXiv:2106.09685}, 2021.

\bibitem[Huang et~al.(2020)Huang, Qiu, and Yuan]{magnetic_tile}
Yibin Huang, Congying Qiu, and Kui Yuan.
\newblock Surface defect saliency of magnetic tile.
\newblock \emph{The Visual Computer}, 36:\penalty0 85--96, 2020.

\bibitem[Inoue et~al.(2018)Inoue, Furuta, Yamasaki, and Aizawa]{comic}
Naoto Inoue, Ryosuke Furuta, Toshihiko Yamasaki, and Kiyoharu Aizawa.
\newblock Cross-domain weakly-supervised object detection through progressive domain adaptation.
\newblock In \emph{Proceedings of the IEEE conference on computer vision and pattern recognition}, pages 5001--5009, 2018.

\bibitem[Jha et~al.(2020)Jha, Smedsrud, Riegler, Halvorsen, de~Lange, Johansen, and Johansen]{jha2020kvasir}
Debesh Jha, Pia~H Smedsrud, Michael~A Riegler, P{\aa}l Halvorsen, Thomas de~Lange, Dag Johansen, and H{\aa}vard~D Johansen.
\newblock Kvasir-seg: A segmented polyp dataset.
\newblock In \emph{MultiMedia Modeling: 26th International Conference, MMM 2020, Daejeon, South Korea, January 5--8, 2020, Proceedings, Part II 26}, pages 451--462. Springer, 2020.

\bibitem[Jin et~al.(2023)Jin, Chollet, Song, and Hu]{jin2023autokeras}
Haifeng Jin, Fran{\c{c}}ois Chollet, Qingquan Song, and Xia Hu.
\newblock Autokeras: An automl library for deep learning.
\newblock \emph{Journal of Machine Learning Research}, 24\penalty0 (6):\penalty0 1--6, 2023.

\bibitem[Khot et~al.(2018)Khot, Sabharwal, and Clark]{Khot2018SciTaiLAT}
Tushar Khot, Ashish Sabharwal, and Peter Clark.
\newblock Scitail: A textual entailment dataset from science question answering.
\newblock In \emph{AAAI Conference on Artificial Intelligence}, 2018.
\newblock URL \url{https://api.semanticscholar.org/CorpusID:24462950}.

\bibitem[Kiela et~al.(2021)Kiela, Firooz, Mohan, Goswami, Singh, Ringshia, and Testuggine]{hateful_meme}
Douwe Kiela, Hamed Firooz, Aravind Mohan, Vedanuj Goswami, Amanpreet Singh, Pratik Ringshia, and Davide Testuggine.
\newblock The hateful memes challenge: Detecting hate speech in multimodal memes, 2021.

\bibitem[Kirillov et~al.(2019)Kirillov, Wu, He, and Girshick]{kirillov2019pointrend}
Alexander Kirillov, Yuxin Wu, Kaiming He, and Ross Girshick.
\newblock {PointRend}: Image segmentation as rendering.
\newblock 2019.

\bibitem[Kirillov et~al.(2023)Kirillov, Mintun, Ravi, Mao, Rolland, Gustafson, Xiao, Whitehead, Berg, Lo, et~al.]{kirillov2023segment}
Alexander Kirillov, Eric Mintun, Nikhila Ravi, Hanzi Mao, Chloe Rolland, Laura Gustafson, Tete Xiao, Spencer Whitehead, Alexander~C Berg, Wan-Yen Lo, et~al.
\newblock Segment anything.
\newblock \emph{arXiv preprint arXiv:2304.02643}, 2023.

\bibitem[Krause et~al.(2013)Krause, Stark, Deng, and Fei-Fei]{stanford_cars}
Jonathan Krause, Michael Stark, Jia Deng, and Li~Fei-Fei.
\newblock 3d object representations for fine-grained categorization.
\newblock In \emph{Proceedings of the IEEE International Conference on Computer Vision Workshops}, pages 379--386, 2013.

\bibitem[Krishna et~al.(2017)Krishna, Zhu, Groth, Johnson, Hata, Kravitz, Chen, Kalantidis, Li, Shamma, et~al.]{visualgenome}
Ranjay Krishna, Yuke Zhu, Oliver Groth, Justin Johnson, Kenji Hata, Joshua Kravitz, Stephanie Chen, Yannis Kalantidis, Li-Jia Li, David~A Shamma, et~al.
\newblock Visual genome: Connecting language and vision using crowdsourced dense image annotations.
\newblock \emph{International journal of computer vision}, 123:\penalty0 32--73, 2017.

\bibitem[Le et~al.(2019)Le, Nguyen, Nie, Tran, and Sugimoto]{le2019anabranch}
Trung-Nghia Le, Tam~V Nguyen, Zhongliang Nie, Minh-Triet Tran, and Akihiro Sugimoto.
\newblock Anabranch network for camouflaged object segmentation.
\newblock \emph{Computer vision and image understanding}, 184:\penalty0 45--56, 2019.

\bibitem[LeDell and Poirier(2020)]{ledell2020h2o}
Erin LeDell and Sebastien Poirier.
\newblock H2o automl: Scalable automatic machine learning.
\newblock In \emph{Proceedings of the AutoML Workshop at ICML}, volume 2020. ICML, 2020.

\bibitem[Lin et~al.(2014)Lin, Maire, Belongie, Hays, Perona, Ramanan, Doll{\'a}r, and Zitnick]{mscoco}
Tsung-Yi Lin, Michael Maire, Serge Belongie, James Hays, Pietro Perona, Deva Ramanan, Piotr Doll{\'a}r, and C~Lawrence Zitnick.
\newblock Microsoft coco: Common objects in context.
\newblock In \emph{Computer Vision--ECCV 2014: 13th European Conference, Zurich, Switzerland, September 6-12, 2014, Proceedings, Part V 13}, pages 740--755. Springer, 2014.

\bibitem[Liu et~al.(2022)Liu, Tam, Muqeeth, Mohta, Huang, Bansal, and Raffel]{liu2022few}
Haokun Liu, Derek Tam, Mohammed Muqeeth, Jay Mohta, Tenghao Huang, Mohit Bansal, and Colin~A Raffel.
\newblock Few-shot parameter-efficient fine-tuning is better and cheaper than in-context learning.
\newblock \emph{Advances in Neural Information Processing Systems}, 35:\penalty0 1950--1965, 2022.

\bibitem[Liu et~al.(2020)Liu, Lian, and Yu]{chest10}
Jingyu Liu, Jie Lian, and Yizhou Yu.
\newblock Chestx-det10: Chest x-ray dataset on detection of thoracic abnormalities, 2020.

\bibitem[Liu et~al.(2019)Liu, Ott, Goyal, Du, Joshi, Chen, Levy, Lewis, Zettlemoyer, and Stoyanov]{liu2019roberta}
Yinhan Liu, Myle Ott, Naman Goyal, Jingfei Du, Mandar Joshi, Danqi Chen, Omer Levy, Mike Lewis, Luke Zettlemoyer, and Veselin Stoyanov.
\newblock Roberta: A robustly optimized bert pretraining approach.
\newblock \emph{arXiv preprint arXiv:1907.11692}, 2019.

\bibitem[Liu et~al.(2021)Liu, Lin, Cao, Hu, Wei, Zhang, Lin, and Guo]{liu2021swin}
Ze~Liu, Yutong Lin, Yue Cao, Han Hu, Yixuan Wei, Zheng Zhang, Stephen Lin, and Baining Guo.
\newblock Swin transformer: Hierarchical vision transformer using shifted windows.
\newblock In \emph{Proceedings of the IEEE/CVF international conference on computer vision}, pages 10012--10022, 2021.

\bibitem[Malo et~al.(2014)Malo, Sinha, Korhonen, Wallenius, and Takala]{financial_news}
Pekka Malo, Ankur Sinha, Pekka Korhonen, Jyrki Wallenius, and Pyry Takala.
\newblock Good debt or bad debt: Detecting semantic orientations in economic texts.
\newblock \emph{Journal of the Association for Information Science and Technology}, 65\penalty0 (4):\penalty0 782--796, 2014.

\bibitem[Micikevicius et~al.(2017)Micikevicius, Narang, Alben, Diamos, Elsen, Garcia, Ginsburg, Houston, Kuchaiev, Venkatesh, et~al.]{micikevicius2017mixed}
Paulius Micikevicius, Sharan Narang, Jonah Alben, Gregory Diamos, Erich Elsen, David Garcia, Boris Ginsburg, Michael Houston, Oleksii Kuchaiev, Ganesh Venkatesh, et~al.
\newblock Mixed precision training.
\newblock \emph{arXiv preprint arXiv:1710.03740}, 2017.

\bibitem[Minaee et~al.(2021)Minaee, Boykov, Porikli, Plaza, Kehtarnavaz, and Terzopoulos]{minaee2021image}
Shervin Minaee, Yuri Boykov, Fatih Porikli, Antonio Plaza, Nasser Kehtarnavaz, and Demetri Terzopoulos.
\newblock Image segmentation using deep learning: A survey.
\newblock \emph{IEEE transactions on pattern analysis and machine intelligence}, 44\penalty0 (7):\penalty0 3523--3542, 2021.

\bibitem[Mishra et~al.(2023)Mishra, Suryavardan, Patwa, Chakraborty, Rani, Reganti, Chadha, Das, Sheth, Chinnakotla, Ekbal, and Kumar]{memotion}
Shreyash Mishra, S~Suryavardan, Parth Patwa, Megha Chakraborty, Anku Rani, Aishwarya Reganti, Aman Chadha, Amitava Das, Amit Sheth, Manoj Chinnakotla, Asif Ekbal, and Srijan Kumar.
\newblock Memotion 3: Dataset on sentiment and emotion analysis of codemixed hindi-english memes, 2023.

\bibitem[Mnih(2013)]{mnih2013machine}
Volodymyr Mnih.
\newblock \emph{Machine learning for aerial image labeling}.
\newblock University of Toronto (Canada), 2013.

\bibitem[M{\"u}ller and Hutter(2021)]{muller2021trivialaugment}
Samuel~G M{\"u}ller and Frank Hutter.
\newblock Trivialaugment: Tuning-free yet state-of-the-art data augmentation.
\newblock In \emph{Proceedings of the IEEE/CVF international conference on computer vision}, pages 774--782, 2021.

\bibitem[Nilsback and Zisserman(2008)]{flower102}
Maria-Elena Nilsback and Andrew Zisserman.
\newblock Automated flower classification over a large number of classes.
\newblock In \emph{2008 Sixth Indian conference on computer vision, graphics \& image processing}, pages 722--729. IEEE, 2008.

\bibitem[Olson and Moore(2016)]{olson2016tpot}
Randal~S Olson and Jason~H Moore.
\newblock Tpot: A tree-based pipeline optimization tool for automating machine learning.
\newblock In \emph{Workshop on automatic machine learning}, pages 66--74. PMLR, 2016.

\bibitem[Pace and Barry(1997)]{cal_house}
R~Kelley Pace and Ronald Barry.
\newblock Sparse spatial autoregressions.
\newblock \emph{Statistics \& Probability Letters}, 33\penalty0 (3):\penalty0 291--297, 1997.

\bibitem[Paleyes et~al.(2022)Paleyes, Urma, and Lawrence]{paleyes2022challenges}
Andrei Paleyes, Raoul-Gabriel Urma, and Neil~D Lawrence.
\newblock Challenges in deploying machine learning: a survey of case studies.
\newblock \emph{ACM computing surveys}, 55\penalty0 (6):\penalty0 1--29, 2022.

\bibitem[pandas~development team(2020)]{reback2020pandas}
The pandas~development team.
\newblock pandas-dev/pandas: Pandas, February 2020.
\newblock URL \url{https://doi.org/10.5281/zenodo.3509134}.

\bibitem[Papafitsoros et~al.(2022)Papafitsoros, Adam, {\v{C}}erm{\'a}k, and Picek]{papafitsoros2022seaturtleid}
Kostas Papafitsoros, Luk{\'a}{\v{s}} Adam, Vojt{\v{e}}ch {\v{C}}erm{\'a}k, and Luk{\'a}{\v{s}} Picek.
\newblock Seaturtleid: A novel long-span dataset highlighting the importance of timestamps in wildlife re-identification.
\newblock \emph{arXiv preprint arXiv:2211.10307}, 2022.

\bibitem[Parisi et~al.(2019)Parisi, Kemker, Part, Kanan, and Wermter]{parisi2019continual}
German~I Parisi, Ronald Kemker, Jose~L Part, Christopher Kanan, and Stefan Wermter.
\newblock Continual lifelong learning with neural networks: A review.
\newblock \emph{Neural networks}, 113:\penalty0 54--71, 2019.

\bibitem[Parkhi et~al.(2012)Parkhi, Vedaldi, Zisserman, and Jawahar]{oxford_pet}
Omkar~M. Parkhi, Andrea Vedaldi, Andrew Zisserman, and C.~V. Jawahar.
\newblock Cats and dogs.
\newblock In \emph{IEEE Conference on Computer Vision and Pattern Recognition}, 2012.

\bibitem[Paszke et~al.(2019)Paszke, Gross, Massa, Lerer, Bradbury, Chanan, Killeen, Lin, Gimelshein, Antiga, et~al.]{paszke2019pytorch}
Adam Paszke, Sam Gross, Francisco Massa, Adam Lerer, James Bradbury, Gregory Chanan, Trevor Killeen, Zeming Lin, Natalia Gimelshein, Luca Antiga, et~al.
\newblock Pytorch: An imperative style, high-performance deep learning library.
\newblock \emph{Advances in neural information processing systems}, 32, 2019.

\bibitem[Radford et~al.(2021)Radford, Kim, Hallacy, Ramesh, Goh, Agarwal, Sastry, Askell, Mishkin, Clark, et~al.]{radford2021learning}
Alec Radford, Jong~Wook Kim, Chris Hallacy, Aditya Ramesh, Gabriel Goh, Sandhini Agarwal, Girish Sastry, Amanda Askell, Pamela Mishkin, Jack Clark, et~al.
\newblock Learning transferable visual models from natural language supervision.
\newblock In \emph{International conference on machine learning}, pages 8748--8763. PMLR, 2021.

\bibitem[Rath(2023)]{Leaf}
Sovit~Ranjan Rath.
\newblock Leaf disease segmentation dataset, 2023.
\newblock URL \url{https://www.kaggle.com/datasets/sovitrath/leaf-disease-segmentation-with-trainvalid-split}.

\bibitem[Reed et~al.(2016)Reed, Akata, Lee, and Schiele]{cub200_text}
Scott Reed, Zeynep Akata, Honglak Lee, and Bernt Schiele.
\newblock Learning deep representations of fine-grained visual descriptions.
\newblock In \emph{Proceedings of the IEEE conference on computer vision and pattern recognition}, pages 49--58, 2016.

\bibitem[Reimers and Gurevych(2019)]{reimers2019sentence}
Nils Reimers and Iryna Gurevych.
\newblock Sentence-bert: Sentence embeddings using siamese bert-networks.
\newblock \emph{arXiv preprint arXiv:1908.10084}, 2019.

\bibitem[Sa et~al.(2016)Sa, Ge, Dayoub, Upcroft, Perez, and McCool]{deepfruits}
Inkyu Sa, Zongyuan Ge, Feras Dayoub, Ben Upcroft, Tristan Perez, and Chris McCool.
\newblock Deepfruits: A fruit detection system using deep neural networks.
\newblock \emph{sensors}, 16\penalty0 (8):\penalty0 1222, 2016.

\bibitem[Sarafianos et~al.(2019)Sarafianos, Xu, and Kakadiaris]{recallmetric}
Nikolaos Sarafianos, Xiang Xu, and Ioannis~A Kakadiaris.
\newblock Adversarial representation learning for text-to-image matching.
\newblock In \emph{Proceedings of the IEEE/CVF international conference on computer vision}, pages 5814--5824, 2019.

\bibitem[Schmarje et~al.(2021)Schmarje, Santarossa, Schr{\"o}der, and Koch]{schmarje2021survey}
Lars Schmarje, Monty Santarossa, Simon-Martin Schr{\"o}der, and Reinhard Koch.
\newblock A survey on semi-, self-and unsupervised learning for image classification.
\newblock \emph{IEEE Access}, 9:\penalty0 82146--82168, 2021.

\bibitem[Schwenk and Li(2018)]{mldoc}
Holger Schwenk and Xian Li.
\newblock A corpus for multilingual document classification in eight languages.
\newblock In Nicoletta Calzolari~(Conference chair), Khalid Choukri, Christopher Cieri, Thierry Declerck, Sara Goggi, Koiti Hasida, Hitoshi Isahara, Bente Maegaard, Joseph Mariani, Hélène Mazo, Asuncion Moreno, Jan Odijk, Stelios Piperidis, and Takenobu Tokunaga, editors, \emph{Proceedings of the Eleventh International Conference on Language Resources and Evaluation (LREC 2018)}, Paris, France, may 2018. European Language Resources Association (ELRA).
\newblock ISBN 979-10-95546-00-9.

\bibitem[Shankar et~al.(2017)Shankar, Nikhil, and Kornel]{quora_duplicate}
Iyer Shankar, Dandekar Nikhil, and Csernai Kornel.
\newblock First quora dataset release: Question pairs.
\newblock Technical report, Quora, 2017.
\newblock URL \url{https://quoradata.quora.com/First-Quora-Dataset-Release-Question-Pairs}.

\bibitem[Shchur et~al.(2023)Shchur, Turkmen, Erickson, Shen, Shirkov, Hu, and Wang]{shchur2023autogluon}
Oleksandr Shchur, Ali~Caner Turkmen, Nick Erickson, Huibin Shen, Alexander Shirkov, Tony Hu, and Bernie Wang.
\newblock Autogluon--timeseries: Automl for probabilistic time series forecasting.
\newblock In \emph{International Conference on Automated Machine Learning}, pages 9--1. PMLR, 2023.

\bibitem[Shorten and Khoshgoftaar(2019)]{shorten2019survey}
Connor Shorten and Taghi~M Khoshgoftaar.
\newblock A survey on image data augmentation for deep learning.
\newblock \emph{Journal of big data}, 6\penalty0 (1):\penalty0 1--48, 2019.

\bibitem[Shorten et~al.(2021)Shorten, Khoshgoftaar, and Furht]{shorten2021text}
Connor Shorten, Taghi~M Khoshgoftaar, and Borko Furht.
\newblock Text data augmentation for deep learning.
\newblock \emph{Journal of big Data}, 8:\penalty0 1--34, 2021.

\bibitem[Singh et~al.(2020)Singh, Jain, Jain, Kayal, Kumawat, and Batra]{plantdoc}
Davinder Singh, Naman Jain, Pranjali Jain, Pratik Kayal, Sudhakar Kumawat, and Nipun Batra.
\newblock Plantdoc: A dataset for visual plant disease detection.
\newblock In \emph{Proceedings of the 7th ACM IKDD CoDS and 25th COMAD}, CoDS COMAD 2020, page 249–253, New York, NY, USA, 2020. Association for Computing Machinery.
\newblock ISBN 9781450377386.
\newblock \doi{10.1145/3371158.3371196}.
\newblock URL \url{https://doi.org/10.1145/3371158.3371196}.

\bibitem[Song et~al.(2016)Song, Xiang, Jegelka, and Savarese]{SOP}
Hyun~Oh Song, Yu~Xiang, Stefanie Jegelka, and Silvio Savarese.
\newblock Deep metric learning via lifted structured feature embedding.
\newblock In \emph{Computer Vision and Pattern Recognition (CVPR)}, 2016.

\bibitem[Sun et~al.(2019)Sun, Xiao, Liu, and Wang]{sun2019deep}
Ke~Sun, Bin Xiao, Dong Liu, and Jingdong Wang.
\newblock Deep high-resolution representation learning for human pose estimation.
\newblock In \emph{Proceedings of the IEEE/CVF conference on computer vision and pattern recognition}, pages 5693--5703, 2019.

\bibitem[Thornton et~al.(2013)Thornton, Hutter, Hoos, and Leyton-Brown]{thornton2013auto}
Chris Thornton, Frank Hutter, Holger~H Hoos, and Kevin Leyton-Brown.
\newblock Auto-weka: Combined selection and hyperparameter optimization of classification algorithms.
\newblock In \emph{Proceedings of the 19th ACM SIGKDD international conference on Knowledge discovery and data mining}, pages 847--855, 2013.

\bibitem[Tschandl et~al.(2018)Tschandl, Rosendahl, and Kittler]{ham10000}
Philipp Tschandl, Cliff Rosendahl, and Harald Kittler.
\newblock The ham10000 dataset, a large collection of multi-source dermatoscopic images of common pigmented skin lesions.
\newblock \emph{Scientific data}, 5\penalty0 (1):\penalty0 1--9, 2018.

\bibitem[Vakhrushev et~al.(2021)Vakhrushev, Ryzhkov, Savchenko, Simakov, Damdinov, and Tuzhilin]{vakhrushev2021lightautoml}
Anton Vakhrushev, Alexander Ryzhkov, Maxim Savchenko, Dmitry Simakov, Rinchin Damdinov, and Alexander Tuzhilin.
\newblock Lightautoml: Automl solution for a large financial services ecosystem.
\newblock \emph{arXiv preprint arXiv:2109.01528}, 2021.

\bibitem[Vanholder(2016)]{vanholder2016efficient}
Han Vanholder.
\newblock Efficient inference with tensorrt.
\newblock In \emph{GPU Technology Conference}, volume~1, 2016.

\bibitem[Vaswani et~al.(2017)Vaswani, Shazeer, Parmar, Uszkoreit, Jones, Gomez, Kaiser, and Polosukhin]{vaswani2017attention}
Ashish Vaswani, Noam Shazeer, Niki Parmar, Jakob Uszkoreit, Llion Jones, Aidan~N Gomez, {\L}ukasz Kaiser, and Illia Polosukhin.
\newblock Attention is all you need.
\newblock \emph{Advances in neural information processing systems}, 30, 2017.

\bibitem[Vicente et~al.(2016)Vicente, Hou, Yu, Hoai, and Samaras]{vicente2016large}
Tom{\'a}s F~Yago Vicente, Le~Hou, Chen-Ping Yu, Minh Hoai, and Dimitris Samaras.
\newblock Large-scale training of shadow detectors with noisily-annotated shadow examples.
\newblock In \emph{Computer Vision--ECCV 2016: 14th European Conference, Amsterdam, The Netherlands, October 11-14, 2016, Proceedings, Part VI 14}, pages 816--832. Springer, 2016.

\bibitem[Wah et~al.(2022)Wah, Branson, Welinder, Perona, and Belongie]{cub200}
Catherine Wah, Steve Branson, Peter Welinder, Pietro Perona, and Serge Belongie.
\newblock Cub-200-2011, Apr 2022.

\bibitem[Wald(1943)]{wald1943tests}
Abraham Wald.
\newblock Tests of statistical hypotheses concerning several parameters when the number of observations is large.
\newblock \emph{Transactions of the American Mathematical society}, 54\penalty0 (3):\penalty0 426--482, 1943.

\bibitem[Wang et~al.(2021{\natexlab{a}})Wang, Wu, Weimer, and Zhu]{wang2021flaml}
Chi Wang, Qingyun Wu, Markus Weimer, and Erkang Zhu.
\newblock Flaml: A fast and lightweight automl library.
\newblock \emph{Proceedings of Machine Learning and Systems}, 3:\penalty0 434--447, 2021{\natexlab{a}}.

\bibitem[Wang et~al.(2021{\natexlab{b}})Wang, Ding, Zhai, Zhang, Tang, Zheng, and Hua]{ipanda50}
Le~Wang, Rizhi Ding, Yuanhao Zhai, Qilin Zhang, Wei Tang, Nanning Zheng, and Gang Hua.
\newblock Giant panda identification.
\newblock \emph{IEEE Transactions on Image Processing}, 30:\penalty0 2837--2849, 2021{\natexlab{b}}.
\newblock \doi{10.1109/TIP.2021.3055627}.

\bibitem[Wightman(2019)]{rw2019timm}
Ross Wightman.
\newblock Pytorch image models.
\newblock \url{https://github.com/rwightman/pytorch-image-models}, 2019.

\bibitem[Williams et~al.(2018)Williams, Nangia, and Bowman]{multinli}
Adina Williams, Nikita Nangia, and Samuel Bowman.
\newblock A broad-coverage challenge corpus for sentence understanding through inference.
\newblock In \emph{Proceedings of the 2018 Conference of the North American Chapter of the Association for Computational Linguistics: Human Language Technologies, Volume 1 (Long Papers)}, pages 1112--1122. Association for Computational Linguistics, 2018.
\newblock URL \url{http://aclweb.org/anthology/N18-1101}.

\bibitem[Wolf et~al.(2020)Wolf, Debut, Sanh, Chaumond, Delangue, Moi, Cistac, Rault, Louf, Funtowicz, et~al.]{wolf2020transformers}
Thomas Wolf, Lysandre Debut, Victor Sanh, Julien Chaumond, Clement Delangue, Anthony Moi, Pierric Cistac, Tim Rault, R{\'e}mi Louf, Morgan Funtowicz, et~al.
\newblock Transformers: State-of-the-art natural language processing.
\newblock In \emph{Proceedings of the 2020 conference on empirical methods in natural language processing: system demonstrations}, pages 38--45, 2020.

\bibitem[Wu et~al.(2019)Wu, Kirillov, Massa, Lo, and Girshick]{detectron2}
Yuxin Wu, Alexander Kirillov, Francisco Massa, Wan-Yen Lo, and Ross Girshick.
\newblock Detectron2.
\newblock \url{https://github.com/facebookresearch/detectron2}, 2019.

\bibitem[Xia et~al.(2018)Xia, Bai, Ding, Zhu, Belongie, Luo, Datcu, Pelillo, and Zhang]{dota}
Gui-Song Xia, Xiang Bai, Jian Ding, Zhen Zhu, Serge Belongie, Jiebo Luo, Mihai Datcu, Marcello Pelillo, and Liangpei Zhang.
\newblock Dota: A large-scale dataset for object detection in aerial images.
\newblock In \emph{Proceedings of the IEEE conference on computer vision and pattern recognition}, pages 3974--3983, 2018.

\bibitem[Xiao et~al.(2017)Xiao, Rasul, and Vollgraf]{fashion_mnist}
Han Xiao, Kashif Rasul, and Roland Vollgraf.
\newblock Fashion-mnist: a novel image dataset for benchmarking machine learning algorithms, 2017.

\bibitem[Xie et~al.(2021)Xie, Wang, Wang, Sun, Xu, Liang, and Luo]{xie2021segmenting}
Enze Xie, Wenjia Wang, Wenhai Wang, Peize Sun, Hang Xu, Ding Liang, and Ping Luo.
\newblock Segmenting transparent object in the wild with transformer.
\newblock \emph{arXiv preprint arXiv:2101.08461}, 2021.

\bibitem[Xu et~al.(2020)Xu, Haider, and Mansour]{multiatis}
Weijia Xu, Batool Haider, and Saab Mansour.
\newblock End-to-end slot alignment and recognition for cross-lingual nlu.
\newblock \emph{arXiv preprint arXiv:2004.14353}, 2020.

\bibitem[Yan et~al.(2018)Yan, Wang, Lu, and Summers]{deeplesion}
Ke~Yan, Xiaosong Wang, Le~Lu, and Ronald~M Summers.
\newblock Deeplesion: automated mining of large-scale lesion annotations and universal lesion detection with deep learning.
\newblock \emph{Journal of medical imaging}, 5\penalty0 (3):\penalty0 036501--036501, 2018.

\bibitem[Yao et~al.(2018)Yao, Wang, Chen, Dai, Li, Tu, Yang, and Yu]{yao2018taking}
Quanming Yao, Mengshuo Wang, Yuqiang Chen, Wenyuan Dai, Yu-Feng Li, Wei-Wei Tu, Qiang Yang, and Yang Yu.
\newblock Taking human out of learning applications: A survey on automated machine learning.
\newblock \emph{arXiv preprint arXiv:1810.13306}, 2018.

\bibitem[Young et~al.(2014)Young, Lai, Hodosh, and Hockenmaier]{flickr30k}
Peter Young, Alice Lai, Micah Hodosh, and Julia Hockenmaier.
\newblock From image descriptions to visual denotations: New similarity metrics for semantic inference over event descriptions.
\newblock \emph{Transactions of the Association for Computational Linguistics}, 2:\penalty0 67--78, 2014.
\newblock \doi{10.1162/tacl_a_00166}.
\newblock URL \url{https://aclanthology.org/Q14-1006}.

\bibitem[Yuan and Wang(2018)]{YuanW18}
Yuhui Yuan and Jingdong Wang.
\newblock Ocnet: Object context network for scene parsing.
\newblock \emph{arXiv preprint arXiv:1809.00916}, 2018.

\bibitem[Yuan et~al.(2020)Yuan, Xie, Chen, and Wang]{YuanXCW20}
Yuhui Yuan, Jingyi Xie, Xilin Chen, and Jingdong Wang.
\newblock Segfix: Model-agnostic boundary refinement for segmentation.
\newblock \emph{arXiv preprint arXiv:2007.04269}, 2020.

\bibitem[Zaken et~al.(2021)Zaken, Ravfogel, and Goldberg]{zaken2021bitfit}
Elad~Ben Zaken, Shauli Ravfogel, and Yoav Goldberg.
\newblock Bitfit: Simple parameter-efficient fine-tuning for transformer-based masked language-models.
\newblock \emph{arXiv preprint arXiv:2106.10199}, 2021.

\bibitem[Zhang et~al.(2022)Zhang, Li, Liu, Zhang, Su, Zhu, Ni, and Shum]{dino}
Hao Zhang, Feng Li, Shilong Liu, Lei Zhang, Hang Su, Jun Zhu, Lionel~M Ni, and Heung-Yeung Shum.
\newblock Dino: Detr with improved denoising anchor boxes for end-to-end object detection.
\newblock \emph{arXiv preprint arXiv:2203.03605}, 2022.

\bibitem[Zhang et~al.(2015)Zhang, Zhao, and LeCun]{ag_news}
Xiang Zhang, Junbo Zhao, and Yann LeCun.
\newblock Character-level convolutional networks for text classification.
\newblock \emph{Advances in neural information processing systems}, 28, 2015.

\bibitem[Zhao et~al.(2019)Zhao, Wang, Yang, and Cai]{zhao2019region}
Shuai Zhao, Yang Wang, Zheng Yang, and Deng Cai.
\newblock Region mutual information loss for semantic segmentation.
\newblock \emph{Advances in Neural Information Processing Systems}, 32, 2019.

\bibitem[Zheng and Casari(2018)]{zheng2018feature}
Alice Zheng and Amanda Casari.
\newblock \emph{Feature engineering for machine learning: principles and techniques for data scientists}.
\newblock " O'Reilly Media, Inc.", 2018.

\bibitem[Zhong et~al.(2024)Zhong, Tang, He, Fang, and Yuan]{convlora}
Zihan Zhong, Zhiqiang Tang, Tong He, Haoyang Fang, and Chun Yuan.
\newblock Convolution meets lora: Parameter efficient finetuning for segment anything model.
\newblock \emph{arXiv preprint arXiv:2401.17868}, 2024.

\bibitem[Zhou et~al.(2022)Zhou, Yu, Xie, Xiao, Anandkumar, Feng, and Alvarez]{fan}
Daquan Zhou, Zhiding Yu, Enze Xie, Chaowei Xiao, Anima Anandkumar, Jiashi Feng, and Jose~M. Alvarez.
\newblock Understanding the robustness in vision transformers.
\newblock In \emph{International Conference on Machine Learning (ICML)}, 2022.

\bibitem[Zimmer et~al.(2021)Zimmer, Lindauer, and Hutter]{zimmer2021auto}
Lucas Zimmer, Marius Lindauer, and Frank Hutter.
\newblock Auto-pytorch: Multi-fidelity metalearning for efficient and robust autodl.
\newblock \emph{IEEE Transactions on Pattern Analysis and Machine Intelligence}, 43\penalty0 (9):\penalty0 3079--3090, 2021.

\bibitem[Z{\"o}ller and Huber(2021)]{zoller2021benchmark}
Marc-Andr{\'e} Z{\"o}ller and Marco~F Huber.
\newblock Benchmark and survey of automated machine learning frameworks.
\newblock \emph{Journal of artificial intelligence research}, 70:\penalty0 409--472, 2021.

\bibitem[Zou et~al.(2023)Zou, Chen, Shi, Guo, and Ye]{zou2023object}
Zhengxia Zou, Keyan Chen, Zhenwei Shi, Yuhong Guo, and Jieping Ye.
\newblock Object detection in 20 years: A survey.
\newblock \emph{Proceedings of the IEEE}, 2023.

\end{thebibliography}

% and for biber / biblatex, use:

% \printbibliography

% supplemental material -- everything hereafter will be suppressed during
% submission time if the hidesupplement option is provided!

\newpage
\section*{Submission Checklist}

\begin{enumerate}
\item For all authors\dots
  \begin{enumerate}
  \item Do the main claims made in the abstract and introduction accurately
    reflect the paper's contributions and scope?
    \answerYes{All claims are supported by the experimental evaluation in Section \ref{sec:exp}.}
  \item Did you describe the limitations of your work?
    \answerYes{See Section \ref{sec:conclusion-future-work}.}
  \item Did you discuss any potential negative societal impacts of your work?
    \answerYes{See Section \ref{sec:broader-impact}.}
  \item Did you read the ethics review guidelines and ensure that your paper
    conforms to them? \url{https://2022.automl.cc/ethics-accessibility/}
    \answerYes{The paper conforms to the guidelines.}
  \end{enumerate}
\item If you ran experiments\dots
  \begin{enumerate}
  \item Did you use the same evaluation protocol for all methods being compared (e.g.,
    same benchmarks, data (sub)sets, available resources)?
    \answerYes{All methods of one task use an identical evaluation protocol.}
  \item Did you specify all the necessary details of your evaluation (e.g., data splits,
    pre-processing, search spaces, hyperparameter tuning)?
    \answerYes{See the appendix for details.}
  \item Did you repeat your experiments (e.g., across multiple random seeds or splits) to account for the impact of randomness in your methods or data?
    \answerYes{We ran experiments with multiple random seeds on basic tasks and semantic matching. We used a fixed seed for object detection and semantic segmentation experiments due to the high cost and limited budget. Given that AutoMM significantly outperforms the comparative methods on the detection datasets, we don't expect a different seeds can close the gap.}
  \item Did you report the uncertainty of your results (e.g., the variance across random seeds or splits)?
    \answerYes{We include the uncertainty measurements in Tables \ref{tab:basic_main}, \ref{tab:i2i_results}, \ref{tab:t2t_results}, \ref{tab:i2t_results}, and \ref{tab:t2i_results}.}
  \item Did you report the statistical significance of your results?
    \answerYes{See results in Section \ref{sec:exp}.}
  \item Did you use tabular or surrogate benchmarks for in-depth evaluations?
    \answerNo{Tabular-only tasks are out of the scope of this work.}
  \item Did you compare performance over time and describe how you selected the maximum duration?
    \answerNo{We fixed the epoch number for each task.}
  \item Did you include the total amount of compute and the type of resources
    used (e.g., type of \textsc{gpu}s, internal cluster, or cloud provider)?
    \answerYes{See the appendix for details.}
  \item Did you run ablation studies to assess the impact of different
    components of your approach?
    \answerYes{See Section \ref{sec:ablation} for details.}
  \end{enumerate}
\item With respect to the code used to obtain your results\dots
  \begin{enumerate}
\item Did you include the code, data, and instructions needed to reproduce the
    main experimental results, including all requirements (e.g.,
    \texttt{requirements.txt} with explicit versions), random seeds, an instructive
    \texttt{README} with installation, and execution commands (either in the
    supplemental material or as a \textsc{url})?
    \answerYes{See Section \ref{sec:reproducibility} for details.}
  \item Did you include a minimal example to replicate results on a small subset
    of the experiments or on toy data?
    \answerYes{See Section \ref{sec:reproducibility} for details.}
  \item Did you ensure sufficient code quality and documentation so that someone else
    can execute and understand your code?
    \answerYes{The code is properly documented and we made sure that it can be executed in a fresh environment.}
  \item Did you include the raw results of running your experiments with the given
    code, data, and instructions?
    \answerYes{See Section \ref{sec:reproducibility} for details.}
  \item Did you include the code, additional data, and instructions needed to generate
    the figures and tables in your paper based on the raw results?
     \answerNo{We provide the raw data and describe the procedure in the paper, which should make reproducing the results and figures straightforward.}
  \end{enumerate}
\item If you used existing assets (e.g., code, data, models)\dots
  \begin{enumerate}
  \item Did you cite the creators of used assets?
    \answerYes{References for all used datasets and methods are provided in the appendix.}
  \item Did you discuss whether and how consent was obtained from people whose
    data you're using/curating if the license requires it?
    \answerNA{The evaluation was performed using public benchmark datasets.}
  \item Did you discuss whether the data you are using/curating contains
    personally identifiable information or offensive content?
    \answerNA{The evaluation was performed using public benchmark datasets.}
  \end{enumerate}
\item If you created/released new assets (e.g., code, data, models)\dots
  \begin{enumerate}
    \item Did you mention the license of the new assets (e.g., as part of your code submission)?
    \answerNo{This paper does not introduce any new public assets. The AutoGluon library was released under the Apache 2.0 License.}
    \item Did you include the new assets either in the supplemental material or as
    a \textsc{url} (to, e.g., GitHub or Hugging Face)?
    \answerNo{This paper does not introduce any new public assets.}
  \end{enumerate}
\item If you used crowdsourcing or conducted research with human subjects\dots
  \begin{enumerate}
  \item Did you include the full text of instructions given to participants and
    screenshots, if applicable?
    \answerNA{We did not use crowdsourcing or conduct research with human subjects.}
  \item Did you describe any potential participant risks, with links to
    Institutional Review Board (\textsc{irb}) approvals, if applicable?
    \answerNA{We did not use crowdsourcing or conduct research with human subjects.}
  \item Did you include the estimated hourly wage paid to participants and the
    total amount spent on participant compensation?
    \answerNA{We did not use crowdsourcing or conduct research with human subjects.}
  \end{enumerate}
\item If you included theoretical results\dots
  \begin{enumerate}
  \item Did you state the full set of assumptions of all theoretical results?
    \answerNA{The paper contains no theoretical results.}
  \item Did you include complete proofs of all theoretical results?
    \answerNA{The paper contains no theoretical results.}
  \end{enumerate}
\end{enumerate}

\newpage
\appendix
\section{Reproducibility}\label{sec:reproducibility}
We provide comprehensive instructions and scripts to reproduce this paper's results in the following Github repository: \url{https://github.com/tonyhoo/automm-benchmark-paper}. Within this repository, you can find the raw data results as well. Moreover, AutoMM offers a tutorial website (\url{https://auto.gluon.ai/stable/tutorials/multimodal/index.html}) hosting many hands-on tutorials such as:
\begin{itemize}
    \item Image prediction (\url{https://auto.gluon.ai/stable/tutorials/multimodal/image_prediction/beginner_image_cls.html})
    \item Text prediction
    (\url{https://auto.gluon.ai/stable/tutorials/multimodal/text_prediction/beginner_text.html})
    \item Multimodal prediction
    (\url{https://auto.gluon.ai/stable/tutorials/multimodal/multimodal_prediction/beginner_multimodal.html})
    \item Image-to-image semantic matching
    (\url{https://auto.gluon.ai/stable/tutorials/multimodal/semantic_matching/image2image_matching.html})
    \item Text-to-text semantic matching
    (\url{https://auto.gluon.ai/stable/tutorials/multimodal/semantic_matching/text2text_matching.html})
    \item Image--text semantic matching
    (\url{https://auto.gluon.ai/stable/tutorials/multimodal/semantic_matching/image_text_matching.html})
    \item Object detection
    (\url{https://auto.gluon.ai/stable/tutorials/multimodal/object_detection/quick_start/quick_start_coco.html})
    \item Semantic segmentation
    (\url{https://auto.gluon.ai/stable/tutorials/multimodal/image_segmentation/beginner_semantic_seg.html})
\end{itemize}
Each tutorial includes both toy data and explanatory code. Additionally, a Colab link is provided at the beginning of each tutorial, enabling users to execute the tutorial while following along. When running AutoMM tutorials in Colab, it's important to select the GPU accelerator. Note that achieving complete reproducibility across different environments can be challenging due to the numerous dependencies AutoMM relies on. As highlighted in our setup.py file (https://github.com/autogluon/autogluon/blob/master/multimodal/setup.py\#L23-L56), ensuring the reproducibility of each dependency is not always feasible. For instance, PyTorch, one of our crucial dependencies, acknowledges the difficulty in guaranteeing reproducibility across various releases, commits, or platforms. More information on this can be found in their documentation (https://pytorch.org/docs/stable/notes/randomness.html\#reproducibility).

\section{Tutorials}
This section presents an exhaustive list of our hands-on tutorials, with links directing to our official documentation website. Each tutorial comes with a short introduction for better readability.

\subsection{Text Data - Classification/Regression/NER}
\begin{itemize}
    \item \href{https://auto.gluon.ai/stable/tutorials/multimodal/text_prediction/beginner_text.html}{\textcolor{blue}{AutoMM for Text Prediction - Quick Start.}} How to quickly train text prediction models with AutoMM.
    \item \href{https://auto.gluon.ai/stable/tutorials/multimodal/text_prediction/multilingual_text.html}{\textcolor{blue}{AutoMM for Text Prediction - Multilingual Problems.}}  How to use AutoMM to build models for non-English languages.
    \item \href{https://auto.gluon.ai/stable/tutorials/multimodal/text_prediction/ner.html}{\textcolor{blue}{AutoMM for Named Entity Recognition - Quick Start.}} How to use AutoMM for entity extraction from text data.
\end{itemize}

\subsection{Image Data – Classification / Regression}
\begin{itemize}
    \item \href{https://auto.gluon.ai/stable/tutorials/multimodal/image_prediction/beginner_image_cls.html}{\textcolor{blue}{AutoMM for Image Classification - Quick Start.}} How to quickly train image classification models with AutoMM.
    \item \href{https://auto.gluon.ai/stable/tutorials/multimodal/image_prediction/clip_zeroshot.html}{\textcolor{blue}{AutoMM for Zero-shot Image Classification.}} How to enable zero-shot image classification in AutoMM via pretrained CLIP. 
\end{itemize}

\subsection{Multimodal Data – Classification / Regression / NER}
\begin{itemize}
    \item \href{https://auto.gluon.ai/stable/tutorials/multimodal/multimodal_prediction/beginner_multimodal.html}{\textcolor{blue}{AutoMM for Image+Text+Tabular - Quick Start.}} How to use AutoMM to train a model on image, text, categorical, and numeric data. 
    \item \href{https://auto.gluon.ai/stable/tutorials/multimodal/multimodal_prediction/multimodal_text_tabular.html}{\textcolor{blue}{AutoMM for Text+Tabular - Quick Start.}} How to apply AutoMM to multimodal data tables with a mix of text, categorical, and numeric columns. 
    \item \href{https://auto.gluon.ai/stable/tutorials/multimodal/multimodal_prediction/multimodal_ner.html}{\textcolor{blue}{AutoMM for Entity Extraction with Text and Image - Quick Start.}} How to use AutoMM to train a model for multimodal named entity recognition.
\end{itemize}

\subsection{Image/Text Data – Semantic Matching}
\begin{itemize}
    \item \href{https://auto.gluon.ai/stable/tutorials/multimodal/semantic_matching/text2text_matching.html}{\textcolor{blue}{Text-to-text Semantic Matching with AutoMM - Quick Start.}} How to use AutoMM to train models for measuring the similarity of two text items.
    \item \href{https://auto.gluon.ai/stable/tutorials/multimodal/semantic_matching/image2image_matching.html}{\textcolor{blue}{Image-to-image Semantic Matching with AutoMM - Quick Start.}} How to use AutoMM to train models for measuring the similarity of two images.
    \item \href{https://auto.gluon.ai/stable/tutorials/multimodal/semantic_matching/image_text_matching.html}{\textcolor{blue}{Image-Text Semantic Matching with AutoMM - Quick Start.}} How to use AutoMM to train models for matching image and text data.
    \item \href{https://auto.gluon.ai/stable/tutorials/multimodal/semantic_matching/zero_shot_img_txt_matching.html}{\textcolor{blue}{Image-Text Semantic Matching with AutoMM - Zero-shot.}} How to use AutoMM for zero-shot matching of image and text data.
    \item \href{https://auto.gluon.ai/stable/tutorials/multimodal/semantic_matching/text_semantic_search.html}{\textcolor{blue}{Text Semantic Search with AutoMM.}} How to use semantic embeddings to improve search ranking performance.
\end{itemize}

\subsection{Image Data – Object Detection}
\begin{itemize}
    \item \href{https://auto.gluon.ai/stable/tutorials/multimodal/object_detection/quick_start/quick_start_coco.html}{\textcolor{blue}{AutoMM for Object Detection - COCO Format.}} How to use AutoMM to quickly train a object detector on a dataset with the COCO format.
    \item \href{https://auto.gluon.ai/stable/tutorials/multimodal/object_detection/data_preparation/object_detection_with_dataframe.html}{\textcolor{blue}{AutoMM for Object Detection - DataFrame Format.}} How to use AutoMM to quickly train a detector with the  data in the DataFrame format. 
    \item \href{https://auto.gluon.ai/stable/tutorials/multimodal/object_detection/data_preparation/prepare_coco17.html}{\textcolor{blue}{Prepare COCO2017 Dataset.}} How to prepare COCO2017 dataset for object detection.
    \item \href{https://auto.gluon.ai/stable/tutorials/multimodal/object_detection/data_preparation/prepare_voc.html}{\textcolor{blue}{Prepare Pascal VOC Dataset.}} How to prepare Pascal VOC dataset for object detection.
    \item \href{https://auto.gluon.ai/stable/tutorials/multimodal/object_detection/data_preparation/prepare_watercolor.html}{\textcolor{blue}{Prepare Watercolor Dataset.}} How to prepare Watercolor dataset for object detection.
    \item \href{https://auto.gluon.ai/stable/tutorials/multimodal/object_detection/data_preparation/prepare_pothole.html}{\textcolor{blue}{Prepare Pathhole Dataset.}} How to prepare Pathhole dataset for object detection.
    \item \href{https://auto.gluon.ai/stable/tutorials/multimodal/object_detection/data_preparation/voc_to_coco.html}{\textcolor{blue}{Convert VOC Format Dataset to COCO Format.}} How to convert a dataset from the VOC format to the COCO format.
    \item \href{https://auto.gluon.ai/stable/tutorials/multimodal/object_detection/advanced/finetune_coco.html}{\textcolor{blue}{Finetune on COCO Format Dataset with Customized Settings.}} How to quickly customize high quality object detection model with AutoMM on COCO format datasets.
\end{itemize}

\subsection{Image Data – Segmentation}
\begin{itemize}
    \item \href{https://auto.gluon.ai/stable/tutorials/multimodal/image_segmentation/beginner_semantic_seg.html}{\textcolor{blue}{AutoMM for Semantic Segmentation - Quick Start.}} How quickly train semantic segmentation models with AutoMM. 
\end{itemize}

\subsection{Advanced Topics}
\begin{itemize}
    \item \href{https://auto.gluon.ai/stable/tutorials/multimodal/advanced_topics/efficient_finetuning_basic.html}{\textcolor{blue}{Single GPU Billion-scale Model Training via Parameter-Efficient Finetuning.}} How to take advantage of large foundation models with the help of parameter-efficient finetuning.
    \item \href{https://auto.gluon.ai/stable/tutorials/multimodal/advanced_topics/hyperparameter_optimization.html}{\textcolor{blue}{Hyperparameter Optimization in AutoMM.}} How to do hyperparameter optimization in AutoMM.
    \item \href{https://auto.gluon.ai/stable/tutorials/multimodal/advanced_topics/model_distillation.html}{\textcolor{blue}{Knowledge Distillation in AutoMM}} How to do knowledge distillation in AutoMM.
    \item \href{https://auto.gluon.ai/stable/tutorials/multimodal/advanced_topics/customization.html}{\textcolor{blue}{Customize AutoMM.}} How to customize AutoMM configurations.
    \item \href{https://auto.gluon.ai/stable/tutorials/multimodal/advanced_topics/presets.html}{\textcolor{blue}{AutoMM Presets.}} How to use AutoMM presets.
    \item \href{https://auto.gluon.ai/stable/tutorials/multimodal/advanced_topics/few_shot_learning.html}{\textcolor{blue}{Few Shot Learning with AutoMM.}} How to use foundation models + SVM for few shot learning.
    \item \href{https://auto.gluon.ai/stable/tutorials/multimodal/advanced_topics/focal_loss.html}{\textcolor{blue}{Handling Class Imbalance with AutoMM - Focal Loss.}} How to use AutoMM to handle class imbalance.
    \item \href{https://auto.gluon.ai/stable/tutorials/multimodal/advanced_topics/tensorrt.html}{\textcolor{blue}{Faster Prediction with TensorRT.}} How to use TensorRT in accelerating AutoMM model inference.
    \item \href{https://auto.gluon.ai/stable/tutorials/multimodal/advanced_topics/continuous_training.html}{\textcolor{blue}{Continuous Training with AutoMM.}} Different use cases for continuous training with AutoMM.
\end{itemize}

\section{Presets}\label{sec:presets}
In the experiments, AutoMM used the \texttt{best\_quality} preset. Given a problem type, the hyperparameters are fixed, which means different datasets of the same problem type use the same hyperparamters without tuning. The detailed hyperparameters of each problem type are provided in Table~\ref{tab:preset}.

The presets hyperparameters were determined through an offline search on our internal benchmark datasets. During this searching, we conducted random search for each task on the combinations of model backbone (task-specific model pools), feature pooling mode (CLS, mean), batch size (4, 8, 16, 32, 64, 128, 256), learning rate (1e-5, 1e-4, 1e-3), learning rate choice (layerwise decay, two-stage, single-stage), weight decay (1e-2, 1e-3, 1e-4), learning rate schedule (cosine decay, multi-step, polynomial decay, no decay), warmup steps (0.1, 0.2, 0.3), patience (5, 10, 20), validation check internal (0.5, 1), and epochs (10, 20, 30, 40, 50, 60). We used the budget of running 2000 jobs per task.

Based on the searched presets, we find that different tasks generally favor different foundation models. For example, the basic task uses deberta-v3 as the text backbone, but the text-text matching uses all-mpnet-base-v2. Most preset hyperparameters of object detection and semantic segmentation are different from those the basic classfication/regression. For instance, the basic task uses layerwise leaning rate decay, object detection uses two-stage learning rate, and semantic segmentation utilizes a single-stage learning rate. While the presets of semantic matching are similar to the basic task, the learning rate of image-text matching is different from the basic task’s (1e-5 vs. 1e-4).

% For additional information on hyperparameters, please refer to our tutorial available at: \url{https://auto.gluon.ai/stable/tutorials/multimodal/advanced_topics/customization.html}

\begin{table}[]
\resizebox{1.0\textwidth}{!}{
\begin{tabular}{l|c|c|c|c|c|c}
\toprule
                            & \begin{tabular}[c]{@{}c@{}}Multimodal\\ Classification/Regression\\ (best\_quality)\end{tabular} & \begin{tabular}[c]{@{}c@{}}Semantic Matching\\ TTM\\ (best\_quality)\end{tabular} & \begin{tabular}[c]{@{}c@{}}Semantic Matching\\ IIM\\ (best\_quality)\end{tabular} & \begin{tabular}[c]{@{}c@{}}Semantic Matching\\ ITM\\ (best\_quality)\end{tabular} & \begin{tabular}[c]{@{}c@{}}Object Detection\\ (best\_quality)\end{tabular} & \begin{tabular}[c]{@{}c@{}}Semantic Segmentation\\ (best\_quality)\end{tabular} \\ \hline
text model                  & deberta-v3-base                                                                                  & \textbackslash{}                                                                  & \textbackslash{}                                                                  & \textbackslash{}                                                                  & \textbackslash{}                                                           & \textbackslash{}                                                                \\ \hline
image model                 & swin\_large                                                                                      & all-mpnet-base-v2                                                                 & swin\_large                                                                       & \textbackslash{}                                                                  & \textbackslash{}                                                           & \textbackslash{}                                                                \\ \hline
tabular model               & ft\_transformer                                                                                  & \textbackslash{}                                                                  & \textbackslash{}                                                                  & \textbackslash{}                                                                  & \textbackslash{}                                                           & \textbackslash{}                                                                \\ \hline

task specific model         & \textbackslash{}                                                                                 & \textbackslash{}                                                                  & \textbackslash{}                                                                  & clip-vit-L14-336                                                                  & dino-5scale\_swin-l                                                        & sam-vit-huge                                                                    \\ \hline
pooling\_mode               & cls                                                                                              & mean                                                                              & \textbackslash{}                                                                  & \textbackslash{}                                                                  & \textbackslash{}                                                           & \textbackslash{}                                                                \\ \hline
batch\_size                 & 128                                                                                              & 128                                                                               & 128                                                                               & 128                                                                               & 32                                                                         & 4                                                                               \\ \hline
precision                   & 16-mixed                                                                                         & 16-mixed                                                                          & 16-mixed                                                                          & 16-mixed                                                                          & 16-mixed                                                                   & 16-mixed                                                                        \\ \hline
learning\_rate              & 1e-4                                                                                             & 1e-4                                                                              & 1e-4                                                                              & 1e-5                                                                              & 1e-4                                                                       & 1e-4                                                                            \\ \hline
lr\_choice                  & layerwise\_decay                                                                                 & layerwise\_decay                                                                  & layerwise\_decay                                                                  & layerwise\_decay                                                                  & two-stage                                                                  & single\_stage                                                                   \\ \hline
layerwise\_lr\_decay                   & 0.9                                                                                              & 0.9                                                                               & 0.9                                                                               & 0.9                                                                               & \textbackslash{}                                                           & \textbackslash{}                                                                \\ \hline
low\_lr\_layers             & \textbackslash{}                                                                                 & \textbackslash{}                                                                  & \textbackslash{}                                                                  & \textbackslash{}                                                                  & {[}"backbone"{]}                                                           & \textbackslash{}                                                                \\ \hline
lr\_multiplier              & \textbackslash{}                                                                                 & \textbackslash{}                                                                  & \textbackslash{}                                                                  & \textbackslash{}                                                                  & 0.1                                                                        & \textbackslash{}                                                                \\ \hline
weight\_decay               & 0.001                                                                                            & 0.001                                                                             & 0.001                                                                             & 0.001                                                                             & 0.0001                                                                     & 0.0001                                                                           \\ \hline
gradient\_clip\_val         & 1                                                                                                & 1                                                                                 & 1                                                                                 & 1                                                                                 & 0.1                                                                        & 1                                                                               \\ \hline
lr\_schedule                & cosine\_decay                                                                                    & cosine\_decay                                                                     & cosine\_decay                                                                     & cosine\_decay                                                                     & multi-step                                                                 & polynomial\_decay                                                               \\ \hline
lr\_steps                   & \textbackslash{}                                                                                 & \textbackslash{}                                                                  & \textbackslash{}                                                                  & \textbackslash{}                                                                  & {[}30, 55{]}                                                               & \textbackslash{}                                                                \\ \hline
warmup\_steps               & 0.1                                                                                              & 0.1                                                                               & 0.1                                                                               & 0.1                                                                               & 0.                                                                         & 0.                                                                              \\ \hline
patience                    & 10                                                                                               & 10                                                                                & 10                                                                                & 10                                                                                & 20                                                                         & 10                                                                              \\ \hline
val\_check\_interval        & 0.5                                                                                              & 0.5                                                                               & 0.5                                                                               & 0.5                                                                               & 1.0                                                                        & 1.0                                                                             \\ \hline
check\_val\_every\_n\_epoch & 1                                                                                                & 1                                                                                 & 1                                                                                 & 1                                                                                 & 1                                                                          & 1                                                                               \\ \hline
max\_epochs                 & 10                                                                                               & 10                                                                                & 10                                                                                & 10                                                                                & 60                                                                         & 30                                                                              \\ \hline
efficient\_finetune         & \textbackslash{}                                                                                 & \textbackslash{}                                                                  & \textbackslash{}                                                                  & \textbackslash{}                                                                  & \textbackslash{}                                                           & lora (r=32,a=32)                                                                 \\
\bottomrule
\end{tabular}
}
\caption{Preset configuration of AutoMM for each problem type.}
\label{tab:preset}
\end{table}

\begin{table}[]
\resizebox{1.0\textwidth}{!}{
\centering
\begin{tabular}{l|c|c|cccc|c}
\toprule
Dataset    & Domain         & Problem Type              & \#Train & \#Val & \#Test & \#Category & Task Description                                    \\ \hline
{fashion\_moist}          & {Fashion}               & {Multiclass} & 60000   & 0     & 10000  & {10}  & {Identify fashion product categories}           \\ \hline 
{food101}                 & {Food}                  & {Multiclass} & 75750   & 0     & 25250  & {101} & {Identify food categories}                      \\ \hline 
{Stanford\_cars}          & {Automotive}            & {Multiclass} & 8144    & 0     & 8041   & {196} & {Identify car models}                           \\ \hline 
{magnetic\_tile\_defects} & {Industrial}            & {Multiclass} & 1008    & 0     & 336    & {6}   & {Identify defects in tiles}                     \\ \hline 
{European\_flood\_depth}  & {Environmental Science} & {Binary}     & 3153    & 0     & 557    & {2}   & {Identify flood types}                          \\ \hline 
{Oxford\_flowers}         & {Botany}                & {Multiclass} & 1020    & 0     & 6149   & {102} & {Identify flower types}                         \\ \hline 
{OxfordIIITPet}           & {Veterinary}            & {Multiclass} & 4436    & 1478  & 1479   & {37}  & {Identify pet breeds}                           \\ \hline 
{CD18\_cellphone}         & {Consumer Product}      & {Regression} & 2532    & 0     & 633    & {NA}  & {Predict cellphone price}                       \\ \hline 
{HAM10000}                & {Medical}               & {Multiclass} & 10015   & 0     & 1512   & {7}   & {Identify dermatological disease types}         \\ \hline 
{hateful\_meme}           & {Social Media}          & {Binary}     & 6800    & 0     & 1700   & {2}   & {Detect harmful content}                        \\ \hline 
{petfinder}               & {Animal Welfare}        & {Multiclass} & 11994   & 0     & 2999   & {5}   & {Predict adoption speed}                        \\ \hline 
{memotion}                & {Social Media}          & {Multiclass} & 5593    & 0     & 1399   & {5}   & {Categorize sentiment}                          \\ \hline 
{financial\_news}         & {Media}                 & {Multiclass} & 3876    & 0     & 969    & {3}   & {Categorize sentiment}                          \\ \hline 
{MLDoc-11000}             & {Information Retrieval} & {Multiclass} & 9777    & 1223  & 4000   & {4}   & {Document classification}                       \\ \hline 
{gnad10}                  & {Social Media}          & {Multiclass} & 8228    & 1017  & 1028   & {9}   & {German news articles categorization}           \\ \hline 
{MultiATIS-5000}          & {Travel}                & {Multiclass} & 4424    & 554   & 891    & {17}  & {Intent recognition}                            \\ \hline 
{fb\_dialog}              & {Social Media}          & {Multiclass} & 25288   & 3162  & 7799   & {12}  & {Intent recognition}                            \\ \hline 
{SNIPS}                   & {Technology}            & {Multiclass} & 13084   & 700   & 700    & {7}   & {Intent recognition}                            \\ \hline 
{ag\_news}                & {Media}                 & {Multiclass} & 106666  & 13334 & 7600   & {4}   & {Identify news topic}                           \\ \hline 
{airbnb\_melbourn}        & {Real Estate}           & {Multiclass} & 18316   & 0     & 4579   & {10}  & {Predict price label}                           \\ \hline 
{kick\_start\_funding}    & {Business}              & {Binary}     & 86502   & 0     & 21626  & {2}   & {Predict crowdfunding campaign’s success}       \\ \hline 
{cloth\_review}           & {Fashion}               & {Multiclass} & 18788   & 0     & 4698   & {5}   & {Categorize sentiment}                          \\ \hline 
{news\_popularity}        & {Media}                 & {Regression} & 24007   & 0     & 6002   & {NA}  & {Predict the number of shares of news articles} \\ \hline 
{California\_house}       & {Real Estate}           & {Regression} & 37951   & 0     & 9488   & {NA}  & {Predict house prices}  \\
\bottomrule
\end{tabular}
}
\caption{Basic Task Datasets}\label{table:basic_data}
\end{table}

\section{Classification and Regression}
\subsection{Dataset Details}
We have included datasets of a wide range of domains to evaluate on the classification and regression tasks, and summarized in Table \ref{table:basic_data}.

\noindent\textbf{Text Classification and Natural Language Understanding.} We include datasets like \texttt{MLDoc-11000}\citep{mldoc} and \texttt{ag\_news}\citep{ag_news} for document and news classification tasks. \texttt{MLDoc-11000} is geared towards multilingual document classification, whereas \texttt{ag\_news} focuses on categorizing news articles. The \texttt{MultiATIS-5000}\citep{multiatis} dataset is used for understanding user input in the context of the Air Travel Information System. \texttt{SNIPS}\citep{snips} is another dataset aimed at natural language understanding, tailored for an AI assistant's conversational understanding. The \texttt{cloth\_review}\citep{cloth_review} dataset provides a basis for sentiment analysis in customer reviews for clothing items, and the \texttt{financial\_news}\citep{financial_news} dataset is used for sentiment analysis within the financial domain.

\noindent\textbf{Image Classification.} For visual recognition tasks, we employ datasets like \texttt{OxfordIIITPet}\citep{oxford_pet}, \texttt{fashion\_mnist}\citep{fashion_mnist}, \texttt{food101}\citep{food101}, and \texttt{stanfordcars}\citep{stanford_cars}. \texttt{OxfordIIITPet}\citep{oxford_pet} contains images for pet breed identification, \texttt{fashion\_mnist} for clothing articles, \texttt{food101} for various food categories, and \texttt{stanfordcars} for automobile models. These datasets vary in complexity and are standard benchmarks in the field of computer vision.

\noindent\textbf{E-Commerce and Housing.} The \texttt{airbnb}\citep{airbnb_melbourne} dataset is included for predicting housing prices and could be used for recommendation systems within the housing rental market. Similarly, the \texttt{cal\_house}\citep{cal_house} dataset provides data for California housing price prediction, which is a classic regression problem.

\noindent\textbf{Medical Imagery.} The \text{ham10000}\citep{ham10000} dataset is integral for medical image analysis, focusing on skin lesion classification for disease detection. 

\noindent\textbf{Defect Detection.} We utilize  
\texttt{magnetictiledefects}\citep{magnetic_tile} dataset for defect detection in manufacturing processes. This dataset is used for automating quality control in an industrial setting.

\noindent\textbf{Sentiment Analysis and Memetics.} The \texttt{hateful\_meme}\citep{hateful_meme} and \texttt{memotion}\citep{memotion} datasets are curated for detecting sentiment and emotion in memes, which is an emerging field in natural language processing and computer vision.

\noindent\textbf{Others.} Additional datasets such as \texttt{petfinder}\citep{petfinder} are used for pet adoption services, and \texttt{news\_popularity}\citep{news_popularity} for predicting the popularity of news articles, demonstrating the varied applications of machine learning across different sectors.

\subsection{Baseline Methods}
We considered several other AutoML frameworks as baselines, e.g. Auto-Sklearn, Auto-Keras, and H2O AutoML.  However, among these, only Auto-Keras supports multimodal input.

\subsection{Metrics}
The metrics being used depend on the specific tasks. 

For regression problems with continuous labels, we use the coefficient of determination, commonly known as R-squared (R²):

\[ R^2 = 1 - \frac{\sum_{i=1}^{n} (y_i - \hat{y}_i)^2}{\sum_{i=1}^{n} (y_i - \bar{y})^2}, \]
where $y_i$ is the value of the label, $\hat{y}_i$ is the predicted value from the regression model, and $\bar{y}$ is the mean of the actual values $y_i$ over all $n$ observations.

For binary classification tasks, we use the F1 score (\(f1\)), which is the harmonic mean of precision and recall:

\[
\mathrm{F1} = 2 \times \frac{\text{precision} \times \text{recall}}{\text{precision} + \text{recall}},
\]
where \[\text{precision} = \frac{\text{True Positive}}{\text{True Positive} + \text{False Positive}} \quad\text{and}\quad\text{recall} = \frac{\text{True Positive}}{\text{True Positive} + \text{False Positive}}\].

For multiclass classification tasks, we use the weighted F1 score (\(f1_{\text{weighted}}\)), which is a weighted average of the F1 score for each class, taking into account label imbalance:

\[
F1_{\text{weighted}} = \sum_{i=1}^{C} w_i \cdot f1_i
\]
where \(C\) is the number of classes, \(\mathrm{F1}_i\) is the F1 score for class \(i\), \(w_i\) is the proportion of true instances for class \(i\) in the dataset.

All reported metrics are computed on a holdout test set not used for training or hyperparameter selection. Some of the datasets include a test set while others do not. In the cases where they do not, we randomly select 20\% of the training data as the test set.

\subsection{Setup}
We used AutoMM's \texttt{best\_quality} presets for all datasets. Since there is no presets provided in Auto-Keras, we adopted the default configuration as suggested by their tutorial. 

We used AWS EC2 g4dn.12xlarge instances equipped with 4 NVIDIA T4 GPUs for all experiment runs. The total GPU time used for training is about 852 hours, which includes 5 repeats with ablation studies on 24 datasets.

\subsection{Ablation Studies}\label{sec:ablation}
Given one pre-trained model, different fine-tuning techniques may result in performance variance. To tease apart the effects of various tricks used in fine-tuning, we conducted extensive ablation studies. AutoMM \texttt{best\_quality} preset have 5 tricks applied (\emph{cosine\_decay}, \emph{gradient\_clip}, \emph{greedy\_soup}, \emph{layerwise\_lr\_decay} and \emph{weight\_decay}). We conducted the experiment by adding each of the 5 tricks to the baseline separately with results presented in columns "\emph{+ trick\_name}", and also by applying all tricks to the baseline ("\emph{+ all}"). The results are shown in Table~\ref{tab:ablations}. Our findings indicate significant performance improvements when employing greedy soup ("+ greedy\_soup") or incorporating all fine-tuning enhancements ("+ all") compared to the baseline model without any modifications ("AutoMM\_base"). Notably, the highest win-rate observed compared across all configurations in our study was achieved with the "+ all" configuration, recording a win-rate of 0.625 (15/24), which forms the basis for our "best\_quality" preset.

\begin{table}[]
\resizebox{1.0\textwidth}{!}{
\begin{tabular}{l|ccccccc}
\toprule
Task                & AutoMM\_base & + cosine\_decay & + grad\_clip & + greedy\_soup &  + layerwise\_lr\_decay & + weight\_decay & + all \\\hline
fashion\_mnist $\uparrow$     & 0.947(0.002)   & 0.947(0.002)                 & 0.947(0.002)              & \textbf{0.955(0.001)}          & 0.948(0.001)             & 0.946(0.003)                 & 0.953(0.002)       \\
food101 $\uparrow$            & 0.920(0.003)   & 0.921(0.001)                 & 0.917(0.002)              & 0.935(0.002)          & 0.923(0.003)             & 0.917(0.003)                 & \textbf{0.937(0.001)}       \\
stanfordcars $\uparrow$       & 0.877(0.003)   & 0.879(0.003)                 & 0.871(0.007)              & 0.890(0.004)          & 0.881(0.003)             & 0.879(0.003)                 & \textbf{0.892(0.002)}       \\
magnetictiledefects $\uparrow$ & 0.955(0.010)   & \textbf{0.964(0.008)}                 & 0.956(0.016)              & 0.961(0.007)          & 0.959(0.010)             & 0.961(0.008)                 & 0.956(0.014)       \\
europeanflooddepth $\uparrow$ & 0.785(0.017)   & 0.789(0.014)                 & 0.786(0.014)              & \textbf{0.796(0.005)}          & 0.786(0.010)             & 0.788(0.007)                 & 0.790(0.008)       \\
oxfordflowers $\uparrow$      & 0.991(0.002)   & 0.990(0.002)                 & 0.991(0.003)              & 0.989(0.004)          & 0.991(0.001)             & \textbf{0.992(0.002)}                 & 0.989(0.003)       \\
OxfordIIITPet $\uparrow$      & 0.955(0.005)   & 0.956(0.003)                 & 0.952(0.003)              & \textbf{0.960(0.002)}          & 0.958(0.006)             & 0.956(0.002)                 & 0.958(0.003)       \\
cd18 $\uparrow$               & -0.037(0.506)  & -2.209(4.705)                & -1.368(3.740)             & \textbf{-0.033(0.507)}         & -1.649(3.426)            & -1.827(4.392)                & -1.843(4.477)      \\
ham10000 $\uparrow$           & 0.532(0.042)   & 0.563(0.023)                 & 0.594(0.024)              & 0.538(0.027)          & 0.588(0.017)             & 0.569(0.020)                 & \textbf{0.608(0.014)}       \\
hateful\_meme $\uparrow$      & 0.577(0.018)   & 0.589(0.027)                 & \textbf{0.612(0.023)}              & 0.590(0.020)          & 0.586(0.013)             & 0.599(0.012)                 & 0.596(0.013)       \\
petfinder $\uparrow$          & 0.398(0.008)   & 0.398(0.006)                 & 0.402(0.007)              & 0.397(0.008)          & 0.403(0.006)             & 0.389(0.011)                 & \textbf{0.408(0.006)}       \\
memotion $\uparrow$           & 0.446(0.014)   & 0.455(0.014)                 & 0.458(0.021)              & 0.449(0.018)          & 0.457(0.014)             & 0.457(0.020)                 & \textbf{0.467(0.013)}       \\
financial\_news $\uparrow$    & 0.868(0.011)   & 0.866(0.013)                 & 0.863(0.004)              & 0.870(0.012)          & 0.868(0.011)             & 0.868(0.006)                 & \textbf{0.874(0.010)}       \\
MLDoc-11000 $\uparrow$        & 0.974(0.002)   & 0.973(0.003)                 & 0.973(0.002)              & 0.977(0.001)          & 0.975(0.002)             & 0.973(0.002)                 & \textbf{0.978(0.002)}       \\
gnad10 $\uparrow$             & 0.881(0.006)   & 0.874(0.007)                 & 0.880(0.004)              & 0.889(0.006)          & 0.891(0.004)             & 0.880(0.008)                 & \textbf{0.899(0.006)}       \\
MultiATIS-5000 $\uparrow$     & 0.989(0.001)   & 0.989(0.002)                 & \textbf{0.990(0.001)}              & \textbf{0.990(0.001)}          & 0.987(0.004)             & 0.987(0.005)                 & 0.990(0.003)       \\
fb\_dialog $\uparrow$         & 0.992(0.001)   & 0.992(0.001)                 & 0.991(0.001)              & \textbf{0.993(0.000)}          & 0.992(0.001)             & 0.991(0.001)                 & 0.992(0.001)       \\
SNIPS $\uparrow$              & 0.989(0.003)   & 0.986(0.003)                 & 0.985(0.001)              & 0.988(0.003)          & 0.987(0.005)             & 0.989(0.002)                 & \textbf{0.990(0.002)}       \\
ag\_news $\uparrow$           & 0.935(0.002)   & 0.938(0.002)                 & 0.938(0.001)              & 0.941(0.002)          & 0.941(0.002)             & 0.937(0.001)                 & \textbf{0.944(0.001)}       \\
airbnb $\uparrow$             & 0.313(0.008)   & 0.309(0.007)                 & 0.323(0.020)              & 0.314(0.009)          & 0.342(0.016)             & 0.305(0.012)                 & \textbf{0.397(0.011)}       \\
kick\_start $\uparrow$        & 0.331(0.166)   & 0.383(0.199)                 & 0.296(0.255)              & 0.415(0.213)          & 0.590(0.009)             & 0.279(0.243)                 & \textbf{0.609(0.005)}       \\
cloth\_review $\uparrow$      & 0.719(0.010)   & 0.717(0.007)                 & 0.722(0.002)              & 0.725(0.012)          & 0.730(0.005)             & 0.716(0.008)                 & \textbf{0.735(0.004)}       \\
news\_popularity $\uparrow$   & 0.008(0.001)   & 0.009(0.003)                 & 0.008(0.001)              & 0.009(0.002)          & 0.012(0.001)             & 0.009(0.003)                 & \textbf{0.014(0.003)}       \\
cal\_house $\uparrow$         & 0.929(0.008)   & 0.931(0.002)                 & 0.937(0.002)              & 0.929(0.003)          & 0.938(0.001)             & 0.929(0.004)                 & \textbf{0.944(0.001)}      \\
\bottomrule
\end{tabular}
}
\caption{Ablation studies for the classification and regression experiments. The numbers indicate the average performance metric (and error bars) for each dataset as we introduce each option. The error bars are estimated using $5$ independent repeats with different random seeds using $1.96 \times \mathrm{std}/\sqrt{\#\text{ of repeats}}$. }\label{tab:ablations}
\end{table}

\begin{table}[]
\resizebox{1.0\textwidth}{!}{
\centering
\begin{tabular}{c|c|cccc|c}
\toprule
Dataset    & Domain                       & \#Train & \#Val & \#Test & \#Category & Task Description                                    \\ \hline
plantdoc   & \multirow{2}{*}{Agriculture} & 8353    & 454   & 457    & 31         & Detection of plant desease detection.               \\
deepfruits &                              & 2552    & 0     & 589    & 7          & Detection of fruits detection.                      \\ \hline
chest10    & \multirow{2}{*}{Medical}     & 6863    & 0     & 1476   & 10         & Detection of Thoracic Abnormalities in chest X-ray. \\
deeplesion &                              & 23785   & 5085  & 5121   & 1          & Detection of lesions in CT images.                  \\ \hline
comic      & Domain Transfer              & 3213    & 0     & 3175   & 6          & Detection of common objects in comic domain.        \\ \hline
dota       & Remote Sencing               & 181746  & 54426 & 236172 & 15         & Detection of objects in aerial images.              \\ \hline
kitti      & Autonomous Driving           & 38076   & 0     & 52457  & 3          & Detection in self driving scenario.                 \\ \hline
thermal    & Infrared                     & 181     & 49    & 27     & 3          & Detection for infrared images.  \\
\bottomrule
\end{tabular}
}
    \caption{ Object Detection Datasets}
    \label{tab:object_detection_dataset}
\end{table}

% Detection
\section{Object Detection}
\subsection{Dataset Details}
Following \citep{aug}, we choose to evaluate the performance on several downstream object detection datasets spanning diverse domains such as agriculture, medical, comic domain transfer, remote sensing, autonomous driving, and infrared imagery. We have excluded simpler datasets while incorporating larger and more intricate ones to ensure a comprehensive evaluation. These datasets are summarized in Table \ref{tab:object_detection_dataset}.

\textbf{Agriculture.} We choose plantdoc~\citep{plantdoc} dataset for plant disease detection and deepfruits~\citep{deepfruits} dataset for fruits detection. Plantdoc has 31 categories, containing 2328 images with 8353 bounding boxes for training and validation, and 239 images with 454 bounding boxes for testing. Deepfruits has 7 categories, containing 457 images with 2552 bounding boxes for training and validation, and 114 images with 589 bounding boxes for testing. 

\textbf{Medical Images.} We choose ChestX-Det10~\citep{chest10} dataset for detection of Thoracic Abnormalities in chest X-ray and deeplesion~\citep{deeplesion} dataset for lesions detection in CT images. ChestX-Det10 has 10 categories, containing 2320 images with 6863 bounding boxes for training and validation, and 459 images with 1476 bounding boxes for testing. Deeplesion has 1 categories, containing 22919 images with 23785 bounding boxes for training, 4889 images with 5085 bounding boxes for validation, and 4927 images with 5121 bounding boxes for testing. 

\textbf{Comic Domain Transfer.} We choose comic~\citep{comic} dataset for detection of common objects in comic domain. Comic has 6 categories, containing 1000 images with 3213 bounding boxes for training and validation, and 1000 images with 3175 bounding boxes for testing. 

\textbf{Remote Sensing.} We choose dota~\citep{dota} dataset for for object detection in aerial images. Dota has 15 categories, containing 9305 images with 181746 bounding boxes for training and validation, and 2955 images with 54426 bounding boxes for testing. 

\textbf{Autonomous Driving.} We choose KITTI~\citep{kitti} dataset for for object detection in the self driving scenario. KITTI has 3 categories, containing 5481 images with 38076 bounding boxes for training and validation, and 7481 images with 52457 bounding boxes for testing. 

\textbf{Infrared Imagery.} We choose Thermal Dogs and People~\citep{roboflow100} dataset for for object detection for infrared images. Thermal Dogs and People dataset has 3 categories, containing 142 images with 181 bounding boxes for training, 41 images with 49 bounding boxes for validation, and 20 images with 27 bounding boxes for testing. 

\subsection{Baseline Methods}
We compare AutoMM with baseline frameworks from both Vertex AI and Nvidia TAO. While Vertex AI offers efficient model development and deployment tools, it may have limitations in pricing flexibility and customization options compared to other solutions. In our experiment, we focused on evaluating Vertex AI only on four datasets using its "Higher accuracy (new)" option due to its high cost, while conducting a thorough comparison with Nvidia TAO using DINO~\citep{dino} with a pretrained FAN-L-Hybrid backbone~\citep{fan} for 60 epochs training. It's important to note that both Vertex AI and Nvidia TAO have specific data requirements, requiring additional preprocessing beyond the common COCO\citep{mscoco} format, while Nvidia TAO additionally requires configuration per dataset.
\subsection{Metrics}
For object detection evaluation, mAP~\citep{mscoco} and AP50~\citep{mscoco} are the key metrics used for assessing both Nvidia Tao and AutoMM models. However, in the case of Vertex AI, only AP50 is reported as the evaluation metric in their service.
\subsection{Raw Results}
We provide the model performance in Table~\ref{tab:det_main}. However, due to the considerable time and computational resources required, we were unable to conduct multiple rounds of experiments for object detection tasks.
\subsection{Setup}
We use a AWS EC2 P4d.24xlarge server with 8x A100 40G GPUs for training and evaluating Nvidia TAO and AutoMM. For Vertex AI, we use their online service API.

% Segmentation
\section{Semantic Segmentation}
\begin{table}[t]
\centering
\resizebox{1.0\textwidth}{!}{
\begin{tabular}{lccccccl}
\toprule
Dataset & Domain & \#Train & \#Val & \#Test & Task Description \\
\hline
Kvasir & Medical & 720 & 180 & 100 & segment abnormal growths within gastrointestinal endoscopic images.\\
CVC-612 & Medical & 440 & 110 & 62 & segment abnormal growths within gastrointestinal endoscopic images.\\
ISIC2017 & Medical & 200 & 600 & 150 & segment various types of skin lesions.\\
CAMO & Natural Images & 3636 & 404 & 250 & segment objects that are concealed within complex backgrounds.\\
SBU & Natural Images & 3677 & 408 & 638 & recognize shadow regions within a scene.\\
Trans10K-v2 & Natural Images & 5000 & 1000 & 4428 & multi-class transparent object segmentation.\\
Leaf & Agriculture & 399 & 99 & 90 & segment individual plant leaf diseases within agricultural images.\\
Road & Remote Sensing & 1107 & 13 & 48 & segment road or street regions within images or video frames.\\

\bottomrule
\end{tabular}
}
\caption{Semantic Segmentation Datasets}
\label{tab:semantic_segmentation_dataset}
\end{table}

\subsection{Dataset Details}
Following \citep{convlora}, we choose several datasets from real-world semantic segmentation scenarios including medical, natural images, agriculture and remote sensing. These datasets are summarized in Table \ref{tab:semantic_segmentation_dataset}.

\textbf{Medical Images.} We choose two polyp segmentation datasets Kvasir~\citep{jha2020kvasir} and CVC-612~\citep{bernal2015wm}, and one skin lesion segmentation dataset ISIC 2017~\citep{codella2018skin}. Kvasir contains 1000 images and CVC-612 includes 612 open-access images. We randomly divide a validation set comprising 20\% of the images from the training set, for validation during training. ISIC 2017 provides 2000 images for training, 150 images for 643
validation and 600 images for testing.

\textbf{Natural Images.} We choose CAMO~\citep{le2019anabranch} for camouflaged object segmentation, SBU~\citep{vicente2016large} for shadow detection, and Trans10K-v2~\citep{xie2021segmenting} dataset for multi-class transparent object segmentation. CAMO provides 1000 images for training and 250 for testing. We train on the combined dataset consists of the training images from COD10K~\citep{fan2020camouflaged} and CAMO for 20 epochs, and test on the three datasets. Additionally, we randomly split 10\% of the images from the training set for validation.  SBU contains 4085 and 638 images for training and testing. We randomly split 10\% of the images from the training set for validation. Trans10K-v2 contains background plus two main categories
divided into 11 fine-grained categories, using 5000, 1000 and 4428 images for training, validation and testing, respectively. 

\textbf{Agriculture.} We use Leaf~\citep{Leaf} dataset for leaf disease segmentation. It contains 498 images for training and 90 images for testing. We randomly split the training images into 80\% for training, 20\% for validation.

\textbf{Remote Sensing.} We choose Massachusetts Roads Dataset~\citep{mnih2013machine} for road segmentation, which contains 1107 images for training, 13 images for validation and 48 images for testing.

\subsection{Baseline Methods}

\textbf{Detectron2}~\citep{detectron2}, developed by Facebook AI Research, is a cutting-edge library offering state-of-the-art detection and segmentation algorithms. We use its supported `SemanticFPN+PointTrend' method as our baseline. PointRend~\citep{kirillov2019pointrend} can be seamlessly integrated into instance and semantic segmentation tasks atop existing state-of-the-art models. And `SemanticFPN+PointTrend' demonstrates superior performance on the Cityscapes~\citep{cordts2016cityscapes} semantic segmentation dataset. 

\textbf{OpenSeg}~\citep{Openseg} is the official PyTorch implementation of the OCNet~\citep{YuanW18} series and SegFix~\citep{YuanXCW20}. We choose its supported `OCR+RMI' method, recognized as the state-of-the-art in their benchmark study. This method employs HRNet~\citep{sun2019deep} as its backbone, ensuring the preservation of high-resolution representations throughout both the image encoding and decoding stages. Moreover, the training loss RMI loss~\citep{zhao2019region}, effectively utilizes region mutual information (RMI) to model dependencies among pixels.

\subsection{Metrics}

We use $S_\alpha$~\citep{fan2017structure} and $E_\phi$~\citep{fan2018enhanced} as our metrics for polyp segmentation and camouflaged object segmentation. These metrics are widely acknowledged within these domains. $S_\alpha$ quantifies the similarity between predictions and ground-truths, while $E_\phi$ provides assessments at both pixel and global levels of similarity. We use Jaccard Index~\citep{codella2018skin} for skin lesion segmentation as the ISBI 2017 challenge ranked methods according to it. We use Balanced Error Rate (BER)~\citep{vicente2016large} for shadow detection, which is a common metric in this area where shadow pixels are considerably less than non-shadow pixels. For leaf segmentation, road segmentation and transparent object detection, we use IoU metric.

\begin{table}[t]\large
    \centering
    \setlength{\tabcolsep}{1.5pt}
    \renewcommand{\arraystretch}{1.5}
    \resizebox{\textwidth}{!}{%
    \begin{tabular}{l|c|c|c|c|c|c|c|c}
        \toprule
        \multirow{2}*{\bf \large Method} &  \multicolumn{3}{c|}{\bf \large Medical} & \multicolumn{3}{c|}{\bf \large Natural Images} & \multicolumn{1}{c|}{\bf \large Agriculture} & \multicolumn{1}{c}{\bf \large Remote Sensing}\\
        \cline{2-9}
        &{ Kvasir} &{CVC-612} & {ISIC 2017} & { CAMO} &   SBU & Trans10K-v2 & \multicolumn{1}{c|}{Leaf} & \multicolumn{1}{c}{ Road} \\

        \midrule
        Detectron2  & 3.3&3.3&10.8&13.7 & 11.3 & 15.6 &1.1& 4.4\\
        OpenSeg  & 3.2&3.2& 6.6 &9.4  & 9.1 & 15.0 &1.3& 2.7  \\
        % MMSeg  & 215.45  & 92.0 & 95.9 & 92.7 & 97.2 & 80.5 & 88.7 & 94.6 & 4.95 &71.5&  80.6 & 53.4 \\
            AutoMM & 5.2 & 5.2 &8.3 & 15.3 & 15.7 & 22.2 & 1.5 & 5.1 \\

        \bottomrule
    \end{tabular}}

    \caption{Training time (hours) of semantic segmentation experiments.}
    \label{tab:seg_training_time}
\end{table}

\subsection{Raw Results}

We provide the model performance in Table~\ref{tab:seg_main}. By leveraging SAM's parameter-efficient fine-tuning, our solution requires only the storage of trainable parameters for each task, unlike other toolboxes that necessitate the storage of entire trainable models for each task during fine-tuning. This enables us to alleviate storage burdens when dealing with multiple segmentation tasks in real-world scenarios, without significantly compromising performance.

Furthermore, our solution demonstrates the capacity to uphold relatively stable performance across datasets from various domains. While Detectron2 and OpenSeg outperform our method on certain datasets (e.g., Trans10K-v2) owing to their larger number of trainable parameters, our approach remains comparable within a performance margin of 6\%. Notably, our method surpasses them by more than 10\% on several other datasets, such as CAMO and Road.

Due to the considerable time and computational resources required, we were unable to conduct multiple rounds of experiments for semantic segmentation tasks. We run all the semantic segmentation experiments on a single NVIDIA V100 32G GPU. The total training time for each experiment is listed in Table~\ref{tab:seg_training_time}.

\section{Semantic Matching}

\begin{table}[t]
\centering
\resizebox{1.0\textwidth}{!}{
\begin{tabular}{lccccccl}
\toprule
Dataset & \#Train & \#Val & \#Test & Pos Ratio & Task & Metric & Task Description \\
\hline
MRPC &  3,261 & 815 & 1,725 &  0.67  & TTM & ROC\_AUC & identify if a sentence from a newswire artile is a paraphrase of another \\
MultiNLI & 160,000 & 40,000  & 9,834 & 0.33 & TTM  & ROC\_AUC & identify if a sentence entails another sentence from different genres  \\
Quora & 323,442 & 40,430 & 40,429 & 0.37 & TTM & ROC\_AUC & identify if a Quora questions is duplicate of another question \\
SciTail & 18,870 & 4,717 & 2,126 & 0.37 & TTM & ROC\_AUC & identify if a sentence entails another sentence in the science domain \\ 
SNLI & 293,282 & 73,321 & 6,605 & 0.50 & TTM & ROC\_AUC & identify if a human-written sentence entails another sentence \\
\hline
SOP & 3,622 & 905 & 2,263 & 0.74 & IIM & ROC\_AUC & identify if two images are from the same online product \\
Airbnb & 1,970 & 493 & 633 & 0.40 & IIM & ROC\_AUC & identify if two images are from the same room in Airbnb \\
CUB200 & 16,151 & 2,020 & 16,515 & 0.27 & IIM & ROC\_AUC & identify if two images are from the same subcategory of birds \\
iPanda50 & 4,000 & 1,000 & 2,000 & 0.30 & IIM & ROC\_AUC & identify if two images are from the same panda individual \\
SeaTurtleID & 4,000 & 1,000 & 2,000 & 0.30 & IIM & ROC\_ACU & identify if two images are from the same turtle individual \\
\hline
CUB200-text & 47,961 & 11,990 & 57,930 & N/A & ITM, TIM & Recall@K & retrieve the most related image/text given the query text/image \\
Flower102-text & 45,850 & 11,462 & 24,570 & N/A & ITM, TIM & Recall@K & retrieve the most related image/text given the query text/image \\
MSCOCO & 14,055 & 3,515 & 7,507 & N/A & ITM, TIM & Recall@K & retrieve the most related image/text given the query text/image \\
Flickr30K & 145,000 & 5,070 & 5,000 & N/A & ITM, TIM & Recall@K & retrieve the most related image/text given the query text/image \\
IPC & 14,579 & 2,490 & 2,492 & N/A & ITM, TIM & Recall@K & retrieve the most related image/text given the query text/image \\
\bottomrule
\end{tabular}
}
\caption{Datasets used in Semantic Matching tasks, Text to Text Matching (TTM), Image to Image Matching (IIM), Text to Image Matching (TIM), Image to Text Matching (ITM). For ITM and TIM, the data only contains positive image-text pairs. For the metric Recall@K, we use K=1, 5, 10, and take the average as the final metric.}
\label{tab:semantic_matching_dataset}
\end{table}

\subsection{Dataset Details}
Table \ref{tab:semantic_matching_dataset} summarizes all the used semantic matching datasets. Below gives detailed descriptions of each dataset.

\textbf{MRPC (Microsoft Research Paraphrase Corpus)} \citep{mrpc} is a corpus consists of 5,801 sentence pairs collected from newswire articles. Each pair is labelled if it is a paraphrase or not by human annotators, which corresponds to matching or not matching in our experiments. We randomly select 20\% of training data as validation data.

\textbf{MultiNLI (Multi-Genre Natural Language Inference)} \citep{multinli} has 433K sentence pairs. It offers ten distinct genres (Face-to-face, Telephone, 9/11, Travel, Letters, Oxford University Press, Slate, Verbatim, Goverment and Fiction) of written and spoken English data. Each sentence pair is labelled as entailment, contradiction, or neutral. In our experiments, we label entailment as matching and contradiction and neutral as not matching. We randomly select 20\% of training data as validation data.

\textbf{Quora Duplicate Questions} \citep{quora_duplicate} contains over 400,000 lines of potential question duplicate pairs. We label duplicate question pairs as matching and non-duplicate pairs as not matching.

\textbf{SciTail} \citep{Khot2018SciTaiLAT} dataset is an entailment dataset created from multiple-choice science exams and web sentences. Each question and the correct answer choice are converted into an assertive statement to form the hypothesis. The premise is obtained from a large text corpus of web sentences. The premise-hypothesis pairs are labelled as entails and neutral, which corresponds to matching and not matching in our experiments. The dataset contains 27,026 examples and we randomly select 20\% of training data as validation data.

\textbf{SNLI (Stanford Natural Language Inference)} consists of 570K sentence-pairs manually labeled as entailment, contradiction, and neutral. In our experiments, we label entailment as matching and contradiction as not matching and discard neutral labels.

\textbf{SOP (Stanford Online Products)} \citep{SOP} contains 12 categories of products. Each category has some products, and each product has several images captured from different views. In the experiments, we consider different views of the same product as matching and images from different products as not matching.

\textbf{Airbnb Duplicate Image} \citep{airbnb} contains interior and exterior house pictures scraped from Airbnb over three cities. Each image in this dataset has at least another image which is a duplicate of the same room. We regard the images of the same room as matching samples and images of different room as not matching.

\textbf{CUB200 (Caltech-UCSD Birds-200)} \citep{cub200} is the most widely-used dataset for fine-grained visual categorization task. It contains 11,788 images of 200 subcategories belonging to birds. \citet{cub200_text} further provided 10 single-sentence descriptions for each image in the dataset, which we name as CUB200-text. In our experiments, we use this dataset for Image to Image Matching, Text to Image Matching and Image to Text Matching tasks. For Image to Image Matching, we randomly select a pair of images and label them as matching pairs if they are from the same category. For Text to Image Matching and Image to Text Matching, we use the image-text pairs from \citet{cub200_text}.

\textbf{iPanda-50} \citep{ipanda50} consists of 6,874 images of 50 giant panda individuals with 49 to 292 images per panda. The iPanda-50 dataset is used for fine-grained panda identification. In our experiments, we use it for Image to Image Matching. Similar to CUB200, we randomly select a pair of images and label them as matching pairs if they are from the same panda.

\begin{table}[t]
    \centering
    \begin{tabular}{l|ccccc}
        \toprule
               & MRPC & MultiNLI & Quora & SciTail & SNLI \\
        \hline
        AutoMM & 87.37 (0.06) & 91.27 (0.04) & 95.77 (0.01) & 97.77 (0.01) & 96.71 (0.01) \\
        Sentence Transformer & 88.89 (0.21) & 91.76 (0.01) & 96.33 (0.01) & 96.93 (0.04) & 96.78 (0.01) \\
        \bottomrule
    \end{tabular}
    \caption{Comparisons on Text to Text Matching datasets. Mean and (Vairiance) are reported.}
    \label{tab:t2t_results}
\end{table}

\begin{table}[t]
    \centering
    \begin{tabular}{l|ccccc}
        \toprule
               & SOP & Airbnb & CUB200 & iPanda & SeaTurtle \\
        \hline
        AutoMM & 97.08 (0.01) & 99.56 (0.00) & 97.83 (0.01) & 82.43 (0.08) & 88.40 (0.33) \\
        Sentence Transformer & 95.97 (0.03) & 98.57 (0.02) & 93.35 (0.01) & 90.40 (0.31) & 89.06 (0.06) \\
        \bottomrule
    \end{tabular}
    \caption{Comparisons on Image to Image Matching datasets. Mean and (Variance) are reported.}
    \label{tab:i2i_results}
\end{table}

\begin{table}[t]
    \centering
    \begin{tabular}{l|ccccc}
        \toprule
               & MSCOCO & Flickr30K & CUB200-text & Flower102-text & IPC \\
        \hline
        AutoMM & 82.44 (0.13) & 92.89 (0.46) & 12.75 (0.12) & 21.31 (0.01) & 91.41 (0.11) \\
        Sentence Transformer & 76.58 (0.04) & 88.31 (0.24) & 15.72 (0.01) & 24.34 (0.02) & 90.12 (0.24) \\
        \bottomrule
    \end{tabular}
    \caption{Comparisons on Text to Image Matching datasets. Mean and (Vairiance) are reported.}
    \label{tab:t2i_results}
\end{table}

\begin{table}[t]
    \centering
    \begin{tabular}{l|ccccc}
        \toprule
               & MSCOCO & Flickr30K & CUB200-text & Flower102-text & IPC \\
        \hline
        AutoMM & 90.29 (0.01) & 98.26 (0.01) & 21.67 (0.24) & 31.44 (0.01) & 91.45 (0.06) \\
        Sentence Transformer & 83.13 (0.58) & 94.90 (0.12) & 25.97 (0.03) & 34.47 (0.11) & 89.73 (0.22) \\
        \bottomrule
    \end{tabular}
    \caption{Comparisons on Image to Text Matching datasets. Mean and (Vairiance) are reported.}
    \label{tab:i2t_results}
\end{table}

\textbf{SeaTurtleID} \citep{papafitsoros2022seaturtleid} is a public large-scale, long-span dataset with sea turtle photographs captured in the wild. It consists of 7774 images of 400 unique individuals collected within 12 years in 1081 encounters. Similar to iPanda-50, we use this dataset for Image to Image Matching, and we randomly select a pair of images and label them as matching pairs if they are from the same turtle.

\textbf{Flower102} \citep{flower102} is a fine-grained image classification dataset consisting of 102 flower categories. Each class consists of between 40 and 258 images. \citet{cub200_text} further provided 10 single-sentence descriptions for each image in the dataset, which we name as Flower102-text. Similar to CUB200, we use this dataset for Image to Image Matching, Text to Image Matching and Image to Text Matching tasks. For Image to Image Matching, we randomly select a pair of images and label them as matching pairs if they are from the same category. For Text to Image Matching and Image to Text Matching, we use the image-text pairs from \citet{cub200_text} 

\textbf{MSCOCO} \citep{mscoco} is a large-scale dataset for object detection, point detection and captioning. It contains 118K training images and 5K validation images. Each image has 5 text descriptions. We used the 5K validation images to build the Image to Text Matching dataset. The train/val/test ratio is roughly 6/1/3.

\textbf{Flickr30K} \citep{flickr30k} is a popular benchmark for sentence-based picture portrayal. The dataset is comprised of 31,783 images that capture people engaged in everyday activities and events. Each image has 5 descriptive captions. This dataset is commonly used as a standard benchmark for Image to Text and Text to Image Matching.

\textbf{IPC (Image Paragraph Captioning)} contains 19,561 images from the Visual Genome dataset \citep{visualgenome}. Each image contains one paragraph describing the image. The training/val/test sets contains 14,575/2,487/2,489 images.

\subsection{Baseline Methods}
We compare AutoMM with Sentence Transformer \citep{reimers2019sentence} on semantic matching tasks. Sentence Transformer could compute embeddings for sentences, paragraphs, and images. The models are based on transformer networks like BERT / RoBERTa / XLM-RoBERTa / CLIP, etc. Text and image are embedded in vector space such that similar text and image are closer and can efficiently be found using cosine similarity. It also allows finetuning these embedding models on the target datasets to achieve maximal performance. For Text to Text Matching, we use its best text embedding model 'all-mpnet-base-v2'. For Image to Image, Image to Text and Text to Image Matching tasks, we use its best image-text embedding model 'CLIP-ViT-L-14'. We use its default hyperparameters for training. We run all the semantic matching experiments on a single NVIDIA A100 40G GPU.

\subsection{Metrics}
For Text to Text and Image to Image Matching tasks, we use Area Under the Receiver Operating Characteristic Curve (ROC\_AUC) as the evaluation metric. For Image to Text and Text to Image Matching tasks, we follow the literature \citep{recallmetric} to use Recall@K (R@K), which is defined as the portion of queries whose ground truth is within the top-K responses. Specifically, we use the average of R@K, where K=1, 5, 10, as the evaluation metric.
\subsection{Raw Results}
We provide the detailed mean and variance of the model performance in Table \ref{tab:t2t_results}-\ref{tab:i2t_results}. We repeat each experiment three times.

\section{Computational Complexity Analysis of AutoMM}

To analyze the computational complexity and performance of AutoMM, we present two tables (Table \ref{tab:computational_complexity_1} and Table \ref{tab:computational_complexity_2}) that showcase various computational metrics for each problem type in an AWS EC2 P4d.24xlarge instance with 8x A100 40G GPUs. The datasets selected for this analysis are Memotion~\cite{memotion}, MRPC~\citep{mrpc}, Airbnb~\citep{airbnb}, CUB200~\citep{cub200}, Comic~\citep{comic}, and Leaf~\citep{Leaf}, with per gpu batch size of 1, 8, 8, 8, 1, and 1 respectively for problem types classification/regression, semantic matching (TTM), semantic matching (IIM), semantic matching (ITM), object detection, and semantic segmentation.

The number of parameters and peak memory usage vary significantly across problem types, indicating the diverse computational requirements of AutoMM. Data preprocessing throughput is generally high, ranging from 3480.7 to 440922.1 samples/s, demonstrating AutoMM's efficiency in handling data preprocessing tasks.

Training and validation throughput values provide insights into the model's efficiency during the learning process, with the highest training throughput observed for semantic matching (ITM) at 489.3 samples/s and the lowest for object detection at 14.1 samples/s. Inference throughput showcases the model's performance during the inference stage, with semantic matching (ITM) having the highest throughput at 9592 samples/s, and semantic segmentation having the lowest at 44.3 samples/s.

Result postprocessing throughput is generally high, ranging from 6523.2 to 183291085 samples/s, indicating AutoMM's efficiency in processing and outputting results across various problem types.

The computational complexity analysis demonstrates the efficiency and scalability of AutoMM across different problem types, highlighting its potential for real-world applications. The tables aim to offer insights into the computational requirements and performance of AutoMM, enabling informed decision-making when applying this approach to specific use cases.

\begin{table}[t]
\centering
\resizebox{1.0\textwidth}{!}{%
\begin{tabular}{l|c|c|c|c|c}
\toprule
\multicolumn{1}{p{2cm}|}{\centering Problem Type}
 & \textbf{\#params (M)} & 
 \multicolumn{1}{|p{2cm}|}{\centering \#Trainable Params (M)}
 & \multicolumn{1}{|p{2cm}|}{\centering Peak Memory (MB)}
  & \multicolumn{1}{|p{2cm}|}{\centering Data Preprocessing Throughput (samples/s)}
  & \multicolumn{1}{|p{2cm}}{\centering Training Throughput (samples/s)} \\ \hline
  \begin{tabular}{@{}c@{}}classification/regression \\ (Image + Text + Tabular)\end{tabular}
 & 183 & 183 & 6032 & 111240.6 & 142.7 \\ \hline
semantic matching (TTM) & 109 & 109 & 8346 & 150947.2 & 362.3 \\ \hline
semantic matching (IIM) & 194 & 194 & 9460 & 110602.5 & 75.8 \\ \hline
semantic matching (ITM) & 427 & 427 & 20042 & 440922.1 & 489.3 \\ \hline
object detection & 218 & 218 & 22398 & 3480.7 & 14.1 \\ \hline
semantic segmentation & 645 & 9 & 25916 & 16224.8 & 22.2 \\
\bottomrule
\end{tabular}
}
\caption{Computational Complexity Analysis of AutoMM (Part 1)}
\label{tab:computational_complexity_1}
\end{table}

\begin{table}[t]
% \resizebox{1.0\textwidth}{!}{
\centering
\begin{tabular}{l|c|c|c|c}
\toprule
\multicolumn{1}{p{2cm}|}{\centering Problem \\Type} & 
 \multicolumn{1}{|p{2cm}|}{\centering Validation Throughput (samples/s)}
 & \multicolumn{1}{|p{2cm}|}{\centering Inference Throughput (samples/s)}
 & \multicolumn{1}{|p{2cm}|}{\centering Inference FPS (single GPU single batch size)}
  & \multicolumn{1}{|p{2cm}}{\centering Result Postprocessing Throughput (samples/s)} \\ \hline
\begin{tabular}{@{}c@{}}classification/regression \\ (Image + Text + Tabular)\end{tabular} & 303.9 & 1165.3 & 35.6 &183291085 \\ \hline
semantic matching (TTM) & 1716.3 & 4072.5 & 41.3 & 229010.1 \\ \hline
semantic matching (IIM) & 164.2 & 131.7 & 16.0 & 79449796.9 \\ \hline
semantic matching (ITM) & 1998.3 & 9592 & 41.5 & 102204.8 \\ \hline
object detection & 66.7 & 62.5 & 8.9 & 6523.2 \\ \hline
semantic segmentation & 57 & 44.3 & 6.9 & 15640442 \\
\bottomrule
\end{tabular}

% }
\caption{Computational Complexity Analysis of AutoMM (Part 2)}
\label{tab:computational_complexity_2}
\end{table}

% \supplemental material can be placed here; this material will be hidden if the
% [hidesupplement] option is provided

\end{document}